\documentclass{article} 

\usepackage[preprint]{neurips_2026}
\usepackage{booktabs} 
\usepackage{graphicx}
\usepackage{subfigure}
\usepackage{multirow}
\usepackage{amsmath}
\usepackage{amssymb}
\usepackage{mathtools}
\usepackage{amsthm}
\usepackage{listings}
\usepackage{titletoc}
\usepackage{wrapfig}
\usepackage{float}

\usepackage{amsmath,amsfonts,bm}

\def\eqref#1{equation~\ref{#1}}

\def\1{\bm{1}}

\DeclareMathAlphabet{\mathsfit}{\encodingdefault}{\sfdefault}{m}{sl}
\SetMathAlphabet{\mathsfit}{bold}{\encodingdefault}{\sfdefault}{bx}{n}

\usepackage{hyperref}
\usepackage{url}


\title{What Matters in On-Policy Distillation? A Perspective on Data Efficiency and Data Selection}

\author{Zhinan Hou\thanks{This work was done during Zhinan’s internship at Meituan.} \\
Tsinghua University \\
\And
Jiaqi Zhang$^\dagger$ \\
Meituan LongCat Team
\And
Xunliang Cai \\
Meituan LongCat Team
\And
Keyou You\thanks{Correspondence to: \texttt{youky@tsinghua.edu.cn}, \texttt{zhangjiaqi39@meituan.com}}\\
Tsinghua University 
}

\begin{document}

\maketitle

\begin{abstract}
On-Policy Distillation (OPD) has emerged as a widely adopted post-training paradigm for enhancing large language models in reasoning domains. However, the data-centric mechanisms in OPD remain relatively underexplored. This paper presents an empirical study of data efficiency and data selection in OPD. We begin by investigating an extreme setting: training OPD on only one example, namely \textbf{1-shot OPD}. Surprisingly, we find that 1-shot OPD is consistently effective across all sampled training examples and harder examples often yield superior performance gain. We next investigate what actually drives the student model’s improvement in the training data. Our analysis reveals that the improvement is not driven by high token entropy, but the longer CoT paths which hard problem naturally generate. Training on longer CoT can help maintain closer alignment with the teacher over a long reasoning horizon, and learn critical thinking patterns usually missing in short CoTs, such as reflection (e.g., \texttt{"Alternatively"}). Based on these insights, we propose a simple data selection method that selects only hard examples for training, where even "unsolvable" examples that completely exceed the teacher's capability can be successfully used. Our experiments conducted on five models ranging from 1.5B to 8B show that training the student model on only 8 selected hard examples matches the performance of the 17K dataset baseline. Code and models will be publicly available.
\end{abstract}

\section{Introduction}
On-policy distillation (OPD) has rapidly emerged as a core technique for large language model (LLM) post-training \citep{lu2025onpolicydistillation}. OPD allows a smaller student model to align its policy with a stronger teacher model by learning directly from on-policy rollouts, enabling the student to effectively inherit complex reasoning capabilities. Consequently, recent pioneering industry efforts, including Qwen3 \citep{yang2025qwen3}, MiMo \citep{xiao2026mimo}, and GLM-5 \citep{zeng2026glm}, have integrated OPD into their post-training pipelines, establishing it as an indispensable complement to supervised fine-tuning (SFT) and reinforcement learning with verified reward (RLVR).

However, while researchers have focused heavily on designing better training algorithms (e.g. EOPD \citep{jin2026entropy} and REOPOLD \citep{ko2026scaling}), the data side of OPD remains relatively underexplored. Specifically, How much data is truly necessary? What data is most effective? And what actually drives the student model's improvement in the training data? Answering these questions is of significant practical value because in many real-world scenarios, high-quality data are extremely scarce or expensive to collect, particularly in highly specialized domains such as medicine and engineering. This bottleneck makes a systematic study of data necessity highly important. 

To answer these questions, we first investigate a special setting where we train OPD on only one example, namely \textbf{1-shot OPD}. Our empirical evaluations demonstrate that 1-shot OPD is consistently effective across all sampled training examples, which is also observed in a concurrent work \citep{fu2026rethinking}. Furthermore, we show that its training is exceptionally stable and the validation performance remains remarkably stable up to even 2,000 steps without overfitting or catastrophic policy collapse. We further analyze why 1-shot OPD works so well and find that the student policy aligns closest with the teacher on structural reasoning tokens (e.g. ``Alternative'', ``Wait'') after distillation. This indicates that the student can successfully learn and generalize the teacher’s reasoning patterns by distilling on only a single training example.

Then, we investigate what kind of data makes OPD most effective. We discover a clear trend: training on harder examples often yields better performance than training on easy or medium ones. Interestingly, even when these examples are so difficult that they completely exceed the capabilities of both the teacher and student models, keeping the training accuracy strictly at 0\%, training on these harder examples still allows the validation accuracy to continuously improve. We further analyze why the example difficulty can influence the OPD and what actually drives the student model's improvement in the training data. We find that this improvement is not driven by high-entropy tokens, but is instead determined by the longer CoT. Specifically, we observe that training with longer CoTs helps the student maintain a closer alignment with the teacher across long horizons. Moreover, these longer paths help the model acquire thinking patterns that are rarely learned from short CoTs, such as self-reflection (e.g., "Alternatively").

Based on these findings, we propose a simple, difficulty-driven data selection method that selects only hard examples for training. We then study how many of these examples are sufficient under this selection. Surprisingly, we find that training a 1.5B student model on only 8 selected hard examples matches the validation accuracy of the full 17K dataset baseline (53.6\% vs 53.7\%). Furthermore, we show that this extreme data efficiency generalizes robustly across diverse model architectures and scales, ranging from 1.5B to 8B parameters.

\section{Preliminaries}
\paragraph{On-Policy Distillation.} OPD computes supervision on trajectories sampled from the current student policy $\pi_\theta$. Given a prompt $x \sim \mathcal{D}_x$, the student samples a response sequence $\hat{y} = (\hat{y}_1, \dots, \hat{y}_N) \sim \pi_\theta(\cdot \mid x)$. Both the student model $\pi_\theta$ and the teacher model $\pi_T$ are evaluated on the student-generated prefixes $\hat{y}_{<t}$, yielding two next-token probability distributions at each step $t$: $p_t(v) \triangleq \pi_\theta(v \mid x, \hat{y}_{<t})$ and $q_t(v) \triangleq \pi_T(v \mid x, \hat{y}_{<t})$ over the vocabulary $v \in \mathcal{V}$.

A standard formulation of OPD minimizes the sequence-level reverse Kullback-Leibler (KL) divergence over the trajectories generated by the student:
\begin{equation}
\mathcal{L}_{\text{OPD}}(\theta) = \mathbb{E}_{x \sim \mathcal{D}_x} \Big[ D_{\text{KL}}(\pi_\theta(\cdot \mid x) \parallel \pi_T(\cdot \mid x)) \Big].
\end{equation}
Using autoregressive factorization, this sequence-level objective can be decomposed into an exact token-level formulation:
\begin{equation}
\mathcal{L}_{\text{OPD}}(\theta) = \mathbb{E}_{x \sim \mathcal{D}_x, \, \hat{y} \sim \pi_\theta(\cdot \mid x)} \left[ \sum_{t=1}^N D_{\text{KL}}(p_t \parallel q_t) \right].
\end{equation}

In practice, existing OPD formulations vary in their supervision granularity for computing this token-level KL, generally categorizing into sampled-token, full-vocabulary, and top-$k$ OPD. Among these, top-$k$ OPD restricts the divergence to a subset of high-probability tokens, significantly reducing memory overhead while preserving dense, multi-token supervision in the student's active generation region. Our experiments are primarily based on top-$k$ OPD.

\paragraph{Top-$k$ OPD.} Top-$k$ OPD restricts the divergence computation to a subset of tokens $S_t \subseteq \mathcal{V}$ at each step. Specifically, we adopt the student top-$k$ variant, which selects the $k$ tokens assigned the highest probabilities under the student distribution, defined as $S_t = \operatorname{TopK}(p_t, k)$. To compute the loss on this subset, we renormalize the student and teacher distributions over $S_t$:
\begin{equation}
\bar{p}_t^{(S_t)}(v) = \frac{p_t(v) \mathbf{1}[v \in S_t]}{\sum_{u \in S_t} p_t(u)}, \quad \bar{q}_t^{(S_t)}(v) = \frac{q_t(v) \mathbf{1}[v \in S_t]}{\sum_{u \in S_t} q_t(u)}.
\end{equation}
Distillation is then performed by minimizing the subset KL divergence $D_{\text{KL}}(\bar{p}_t^{(S_t)} \parallel \bar{q}_t^{(S_t)})$ over the trajectory, yielding the final objective:
\begin{equation}
\mathcal{L}_{\text{OPD}}^{\text{top-}k}(\theta) = \mathbb{E}_{x \sim \mathcal{D}_x, \, \hat{y} \sim \pi_\theta(\cdot \mid x)} \left[ \sum_{t=1}^N D_{\text{KL}}(\bar{p}_t^{(S_t)} \parallel \bar{q}_t^{(S_t)}) \right].
\end{equation}

\section{Pilot Experiments: 1-Shot OPD}
We first study an extreme setting: training a student model on only a single example, termed 1-shot OPD. This special setting serves as a clean baseline to test whether OPD can achieve policy alignment without the influence of data variety. The observation of 1-shot OPD also provides the guidance for our subsequent analyses on what kind of data is suitable and how much data is sufficient.

\subsection{Experimental Setup}
\label{subsec:setup}

\textbf{Models.} We employ DeepSeek-R1-Distill-Qwen-1.5B \citep{guo2025deepseek} as our student model, and JustRL-DeepSeek-1.5B \citep{he2025justrl} as the teacher model. By default, these two models are used for our primary experiments. We also conduct experiments across other model architectures and scales in Section~\ref{subsec:other_models} to validate the generalizability of our findings.

\textbf{Dataset.} Due to computational resource limits, we randomly sample a subset of 1,000 examples from the DAPO-Math-17K dataset \citep{yu2026dapo} to construct an example pool. To ensure reproducibility and consistent identification across all experiments. We simply rank the training examples by difficulty. Specifically, for each example in our pool, we compute the accuracy of both the student and teacher models across 16 rollouts per problem. Let $S_i$ and $T_i$ denote the rollout accuracies of the student and teacher models for the $i$-th problem, respectively. The average accuracy is computed as $A_i = (S_i + T_i)/2$. The training examples are sorted primarily in descending order of $A_i$, and secondarily by their original DAPO-Math-17K index to guarantee a deterministic sequence. The resulting ordered sequence is denoted as $\{\pi_i\}_{i=1}^{1000}$. Based on $A_i$, we further categorize the training examples into three categories: (1) \textbf{Easy} ($A_i > 0.9$), corresponding to indices $\pi_1$ to $\pi_{199}$ (199 examples); (2) \textbf{Medium} ($0.1 \le A_i \le 0.9$), corresponding to indices $\pi_{200}$ to $\pi_{825}$ (626 examples); and (3) \textbf{Hard} ($A_i < 0.1$), corresponding to indices $\pi_{826}$ to $\pi_{1000}$ (175 examples).

\textbf{Training.} OPD is implemented by using the open-source \texttt{verl} \citep{sheng2024hybridflow} framework. Following \citet{li2026rethinking}, the training batch size and mini-batch size are 64, and we sample 8 responses for each prompt. For response generation, we utilize the \texttt{vLLM} engine with a rollout temperature of 1.0 and maximum response length of 7,168 tokens. The per-token reverse KL divergence is computed against only the top-$16$ tokens. We train the OPD in the DAPO-Math-17K for 1 epoch as full dataset baseline (Full-Set OPD), yields 279 training steps. For fair comparisons, we train our 1-shot OPD under 279 training steps by default. To test the domain generalizability of our findings, we also extend our investigation to the code generation domain by distilling on training prompts sampled from the LeetCodeDataset v3.0.1 \citep{xia2025leetcodedataset}. The complete experimental results for the coding domain are presented in Appendix \ref{app:coding}. All experiments are executed on an 8 $\times$ NVIDIA H800 80GB GPU cluster. More training details are included in Appendix \ref{app:opd_training_details}. 

\textbf{Evaluation.} We evaluate the models across six standard mathematical reasoning benchmarks: AIME 2024 \citep{aime}, AIME 2025 \citep{aime}, AMC 2023 \citep{amc}, MATH500 \citep{hendrycks2measuring}, Minerva Math \citep{lewkowycz2022solving}, and OlympiadBench \citep{he2024olympiadbench}. Additionally, we extend our evaluation to non-mathematical domains including LiveCodeBench \citep{jainlivecodebench} and GPQA \citep{rein2023gpqa}. The evaluation results for these non-mathematical datasets are provided in Appendix \ref{app:ood}. For all benchmarks, we sample with a temperature of 0.7 and a maximum output length of 16,384 tokens. Due to the limited sample sizes of AIME 2024, AIME 2025, and AMC 2023, we perform 16 rollouts per problem and report the average accuracy (\textit{mean@16}) to ensure evaluation stability.  For the remaining datasets, we report the average accuracy over 4 rollouts (\textit{mean@4}). 

\begin{figure*}[htbp]
    \centering
    \includegraphics[width=\textwidth]{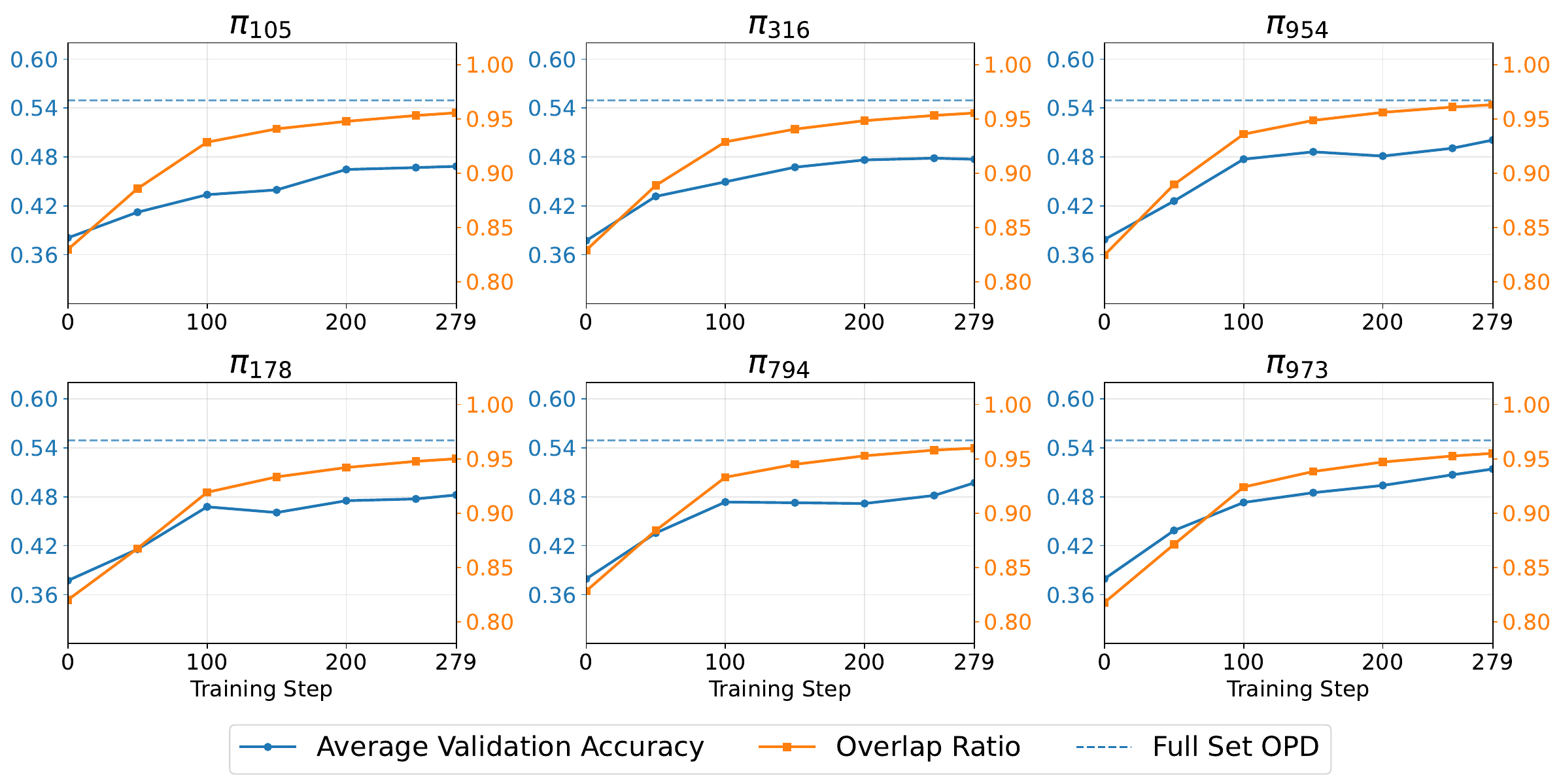}
    \caption{\textbf{Learning Dynamics of 1-shot OPD.} We visualize the training trajectories of six representative examples across three difficulty levels: two easy problems ($\{\pi_{105}\}$ and $\{\pi_{178}\}$, left), two medium problems ($\{\pi_{316}\}$ and $\{\pi_{794}\}$, middle), and two hard problems ($\{\pi_{954}\}$ and $\{\pi_{973}\}$, right). The solid blue curves denote the Average Validation Accuracy (computed across AMC 2023, AIME 2024, and AIME 2025). The solid orange curves with markers track the Overlap Ratio, indicating progressive token-level policy alignment with the teacher model. The horizontal dashed lines represent the full-set baseline (Full-Set OPD) for comparison. }
    \label{fig:training_curves}
\end{figure*}

\subsection{1-Shot OPD is Effective for Many Examples}
\label{subsubsec:one_shot_effectiveness}

\begin{figure}[htbp]
    \centering
    \includegraphics[width=\textwidth]{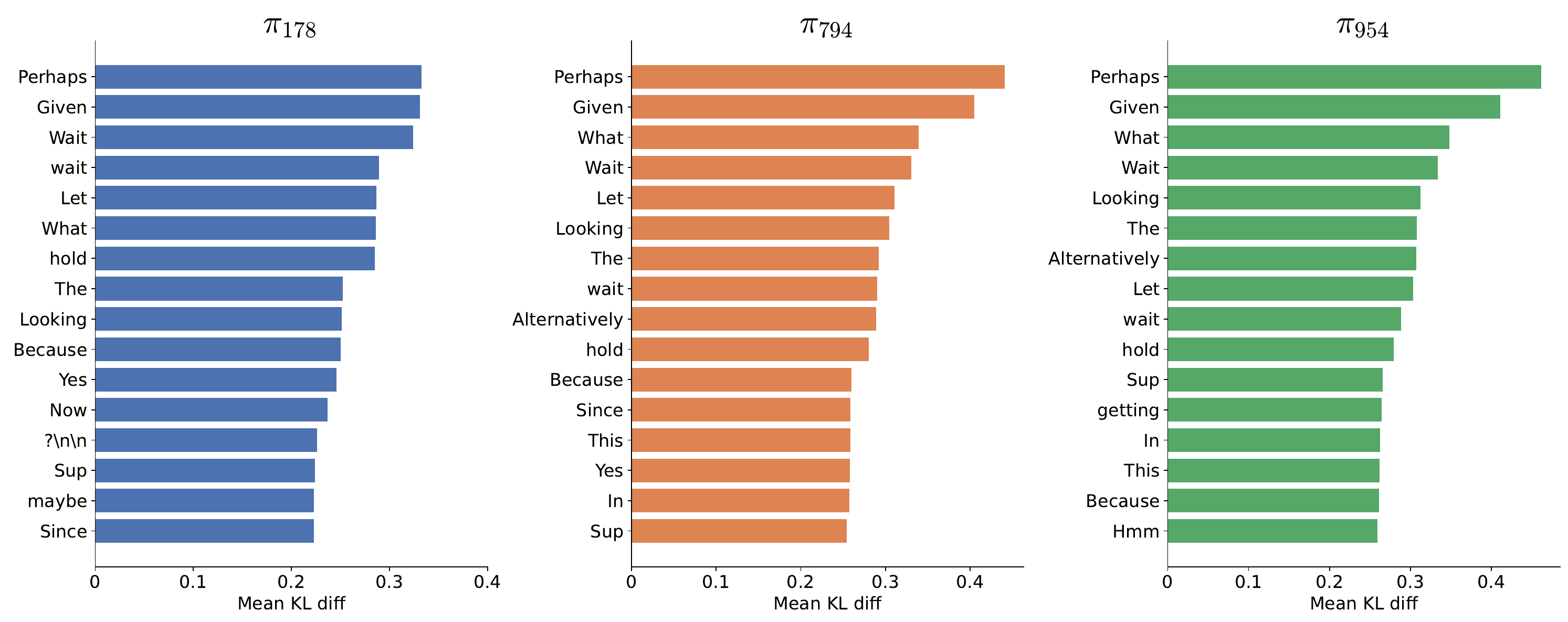}
    \caption{\textbf{Per-Token KL Divergence Reduction post 1-Shot OPD.} We report the top 16 tokens with the highest expected KL reduction after distillation for models trained respectively on $\{\pi_{178}\}$ (left), $\{\pi_{794}\}$ (middle), and $\{\pi_{954}\}$ (right).}
    \label{fig:token_kl_reduction}
\end{figure}

First, we randomly sample 8 training examples from each of the Easy, Medium, and Hard categories and train on them individually. The specific indices of sampled examples are listed in Table \ref{tab:opd_accuracies}. During training, we utilize AMC 2023, AIME 2024, and AIME 2025 as the validation set to monitor performance changes over steps. Consistent with our evaluation, we sample 16 solutions per problem with a temperature of 0.7 and a maximum response length of 16,384 tokens, reporting the average accuracy over the 16 samples as our validation metric. Additionally, we track the overlap ratio \citep{li2026rethinking} to measure policy alignment between the student and the teacher. Specifically, we randomly select 100 examples from the DAPO-Math-17K as a held-out set. At each training checkpoint, we perform 16 independent rollouts for each example. Then we report the average proportion of tokens that appear simultaneously in the top-16 vocabulary distributions of both the student and the teacher as the overlap ratio.

We present the evaluation results across benchmarks in Table \ref{tab:one_shot_benchmark_performance}. Moreover, we select some examples to visualize their training trajectories in Figure~\ref{fig:training_curves}. As illustrated in Figure~\ref{fig:training_curves}, the overlap ratio continuously increases, indicating that the student model is progressively aligning with the teacher model. Moreover, the performance improvement demonstrates the effectiveness of 1-shot OPD. We further perform a prolonged training run of 2,000 steps to observe whether 1-shot OPD suffers from overfitting. As shown in Figure \ref{fig:long_training}, the validation performance plateaus after 300 steps and remains stable up to 2,000 steps without any performance decay or policy collapse.

To understand why 1-shot OPD is so effective, we analyze which tokens shift closest to the teacher after distillation. Specifically, we use the held-out set (100 problems) to generate 16 rollouts per problem. Then we compute the reduction in token-level KL divergence between the student and teacher distributions post-distillation and average this KL reduction over all token IDs. For statistical significance, we discard the rare tokens with an overall occurrence frequency of less than 0.01\%. In Figure \ref{fig:token_kl_reduction}, we report the top 16 tokens with the highest KL reduction for models trained respectively on $\{\pi_{178}\}$, $\{\pi_{794}\}$, and $\{\pi_{954}\}$. The results reveal that the tokens aligning most significantly with the teacher after OPD are predominantly structural reasoning, including reflection (e.g., ``Alternatively''), transitions (e.g., ``Wait'', ``Perhaps''), and deductive reasoning (e.g., ``Because'', ``Since''). This indicates that the student can successfully learn the teacher's reasoning patterns by distilling on only a single training example.

\begin{table}[ht!]
\centering
\caption{\textbf{1-Shot OPD Performance Across Mathematical Reasoning Benchmarks.} We compare the student model DeepSeek-R1-Distill-Qwen-1.5B (Base), the teacher model JustRL-DeepSeek-1.5B (Teacher), and the full-dataset distillation baseline (Full-Set) against our 1-shot OPD models. Notably, 1-shot models trained on hard problems consistently outperform those trained on easy problems, with the best-performing example ($\{\pi_{973}\}$) achieving $51.7\%$ average accuracy, nearly matching the Full-Set baseline ($53.7\%$).}
\label{tab:one_shot_benchmark_performance}
\small
\resizebox{0.8\textwidth}{!}{
\begin{tabular}{lccccccc}
\toprule
\textbf{Model} & \textbf{AIME24} & \textbf{AIME25} & \textbf{AMC23} & \textbf{MATH500} & \textbf{Olympiad} & \textbf{Minerva} & \textbf{Avg} \\
\midrule
Base & 31.7 & 23.0 & 60.7 & 82.6 & 34.1 & 22.4 & 42.4 \\
Teacher & 55.8 & 35.8 & 83.4 & 87.4 & 43.4 & 29.8 & 55.9 \\
Full-Set & 51.9 & 34.0 & 78.9 & 86.7 & 44.3 & 26.6 & 53.7 \\
\midrule
\multicolumn{8}{l}{\textit{Easy Problems}} \\
$\{\pi_{5}\}$ & 39.2 & 28.8 & 69.9 & 85.3 & 39.3 & 25.3 & 48.0 \\
$\{\pi_{20}\}$  & 40.3 & 30.6 & 69.1 & 86.2 & 40.8 & 25.7 & 48.8 \\
$\{\pi_{21}\}$  & 41.8 & 29.0 & 71.1 & 85.9 & 39.7 & 26.2 & 48.9 \\
$\{\pi_{63}\}$  & 42.2 & 30.8 & 71.0 & 86.2 & 39.7 & 25.0 & 49.1 \\
$\{\pi_{84}\}$  & 41.0 & 30.1 & 72.6 & 85.6 & 40.6 & 25.8 & 49.3 \\
$\{\pi_{105}\}$ & 42.5 & 27.7 & 70.3 & 86.0 & 41.1 & 25.0 & 48.8 \\
$\{\pi_{178}\}$ & 41.1 & 30.4 & 71.2 & 86.7 & 40.4 & 25.9 & 49.3 \\
$\{\pi_{192}\}$ & 39.8 & 30.2 & 72.0 & 85.7 & 40.1 & 25.7 & 48.9 \\
\midrule
\multicolumn{8}{l}{\textit{Medium Problems}} \\
$\{\pi_{316}\}$ & 39.8 & 30.0 & 73.3 & 86.4 & 40.6 & 25.7 & 49.3 \\
$\{\pi_{487}\}$ & 41.2 & 29.8 & 72.7 & 85.7 & 40.4 & 25.9 & 49.3 \\
$\{\pi_{530}\}$ & 42.1 & 30.7 & 73.1 & 86.5 & 40.2 & 26.8 & 49.9 \\
$\{\pi_{543}\}$ & 40.4 & 29.4 & 73.0 & 86.4 & 41.3 & 26.9 & 49.6 \\
$\{\pi_{661}\}$ & 43.1 & 31.2 & 73.4 & 86.9 & 41.3 & 26.4 & 50.4 \\
$\{\pi_{763}\}$ & 41.5 & 28.1 & 75.6 & 86.7 & 41.2 & 26.8 & 50.0 \\
$\{\pi_{794}\}$ & 44.2 & 30.8 & 74.2 & 86.5 & 41.6 & 25.7 & 50.5 \\
$\{\pi_{820}\}$ & 41.5 & 29.1 & 73.6 & 84.9 & 40.7 & 25.9 & 49.3 \\
\midrule
\multicolumn{8}{l}{\textit{Hard Problems}} \\
$\{\pi_{874}\}$ & 43.3 & 33.1 & 74.3 & 86.6 & 41.0 & 27.1 & 50.9 \\
$\{\pi_{890}\}$ & 45.2 & 29.6 & 76.0 & 85.6 & 42.4 & 26.5 & 50.9 \\
$\{\pi_{948}\}$ & 45.0 & 29.8 & 75.3 & 86.0 & 41.8 & 27.0 & 50.8 \\
$\{\pi_{954}\}$ & 44.8 & 28.5 & 76.9 & 86.9 & 42.6 & 26.7 & 51.1 \\
$\{\pi_{961}\}$ & 44.4 & 30.6 & 73.2 & 87.2 & 41.4 & 26.7 & 50.6 \\
$\{\pi_{973}\}$ & 47.5 & 31.7 & 75.1 & 87.7 & 41.5 & 26.9 & 51.7 \\
$\{\pi_{987}\}$ & 44.8 & 27.7 & 73.6 & 87.5 & 41.2 & 26.4 & 50.2 \\
$\{\pi_{997}\}$ & 44.0 & 30.2 & 75.8 & 85.6 & 42.1 & 26.8 & 50.8 \\
\bottomrule
\end{tabular}
}
\end{table}

\section{Data Selection for OPD}  \label{sec:selection}
\subsection{Harder Problems Often Lead to Better Performance}
\label{subsubsec:harder_better}

\begin{figure*}[htbp]
    \centering
    \includegraphics[width=\textwidth]{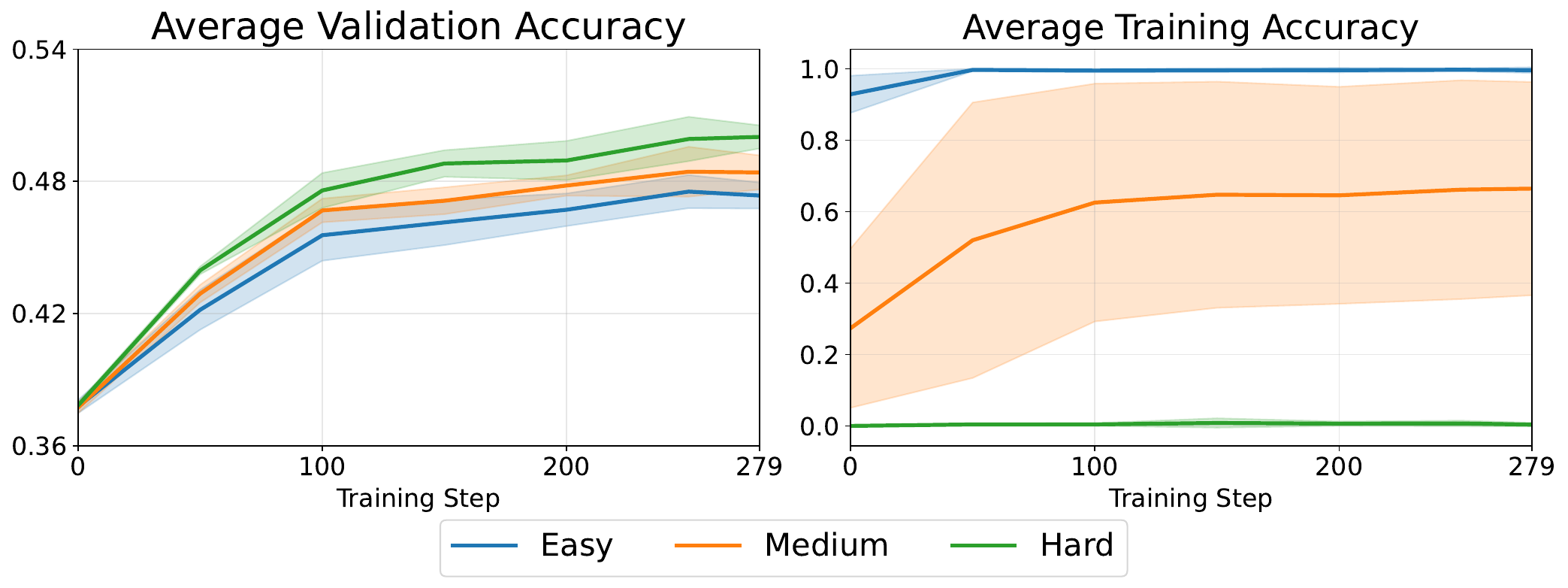}
    \caption{\textbf{Learning Trajectories across Difficulty Levels.} Left: Average Validation Accuracy over training steps. Right: Average Training Accuracy on the selected training example over training steps. Solid lines indicate the mean and shaded areas represent the standard deviation across the examples in each category.}
    \label{fig:difficulty_curves}
\end{figure*}

We further investigate whether different data behave differently in 1-shot OPD. For each category, we record both the average training accuracy and validation accuracy every 50 training steps, and present these trajectories in Figure \ref{fig:difficulty_curves}. Surprisingly, we find that harder training examples often yield better performance. Moreover, we observe that for easy and hard problems, even after their training accuracies saturate at 100\% (for easy tasks) or remain flat at 0\% (for hard tasks), their validation accuracies continuously improve. This behavior contrasts sharply with RL (e.g., GRPO-style algorithms) where training on extreme easy and hard samples is highly ineffective because the computed advantage or gradient updates tend to be zero.

\subsection{Deeper Analysis}
\label{subsec:entropy_vs_cot}
We next explore what underlying properties drive the efficacy of difficult prompts, examining potential factors such as the higher token entropy and the longer CoT reasoning sequences they naturally generate. 

\begin{figure}[t]
    \centering
    \includegraphics[width=\linewidth]{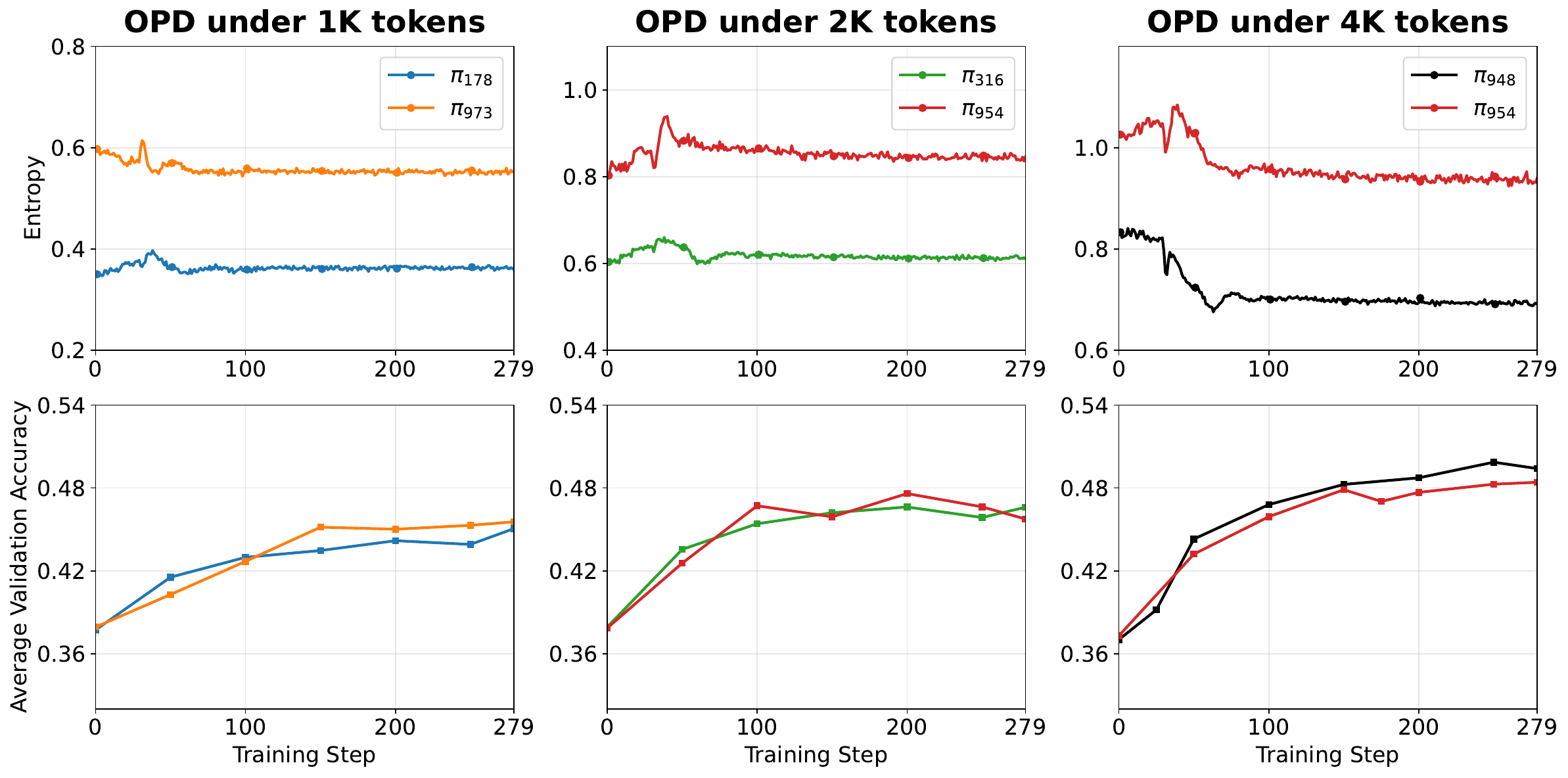}
    \caption{\textbf{Training Dynamics in Length-Constrained Setting}  Training dynamics are presented for problem pairs grouped by difficulty: Easy vs. Hard (1K token limit, left), Medium vs. Hard (2K token limit, middle), and Hard vs. Hard (4K token limit, right). The top plots monitor token entropy, while the bottom plots track validation accuracy.}
    \label{fig:cot_truncation_gap}
\end{figure}

\textbf{Setup.} To isolate the role of token entropy from length related factors, we set up a length-controlled experiment. By adjusting the maximum rollout length, we expect that almost all rollouts generated for some examples will exceed this limit and get cut off. Consequently, training the 1-shot OPD uses almost the same number of tokens for those examples because the length of every response is forced to be the maximum rollout length. Specifically, we conduct three sets of experiments. In each set, we select two problems with different difficulty or entropy, and set an appropriate maximum rollout length such that empirically over 99.5\% of the rollouts for both problems exceed this threshold. The three problem pairs are: 1) Easy problem $\pi_{178}$ and Hard problem $\pi_{973}$ limited to $1024$ tokens, 2) Medium problem $\pi_{316}$ and Hard problem $\pi_{954}$ limited to $2048$ tokens, and 3) two Hard problems with highly different token entropy, $\pi_{948}$ and $\pi_{954}$ limited to $4096$ tokens.

\textbf{Higher Token Entropy Alone Does Not Improve OPD} As illustrated in Figure~\ref{fig:cot_truncation_gap}, under length-controlled settings, we observe no  significant difference in validation performance between the compared pairs, despite their highly divergent token entropy. Specifically, once rollout lengths are restricted to match, the performance gap between the pairs collapses to $0.5\%$ (for $\{\pi_{973}\}$ vs. $\{\pi_{178}\}$) and $-0.9\%$ (for $\{\pi_{954}\}$ vs. $\{\pi_{316}\}$). However, in the default setting (7,168 maximum rollout length) where the student model can leverage the longer CoT sequences generated by harder examples, the harder problems demonstrate a notable advantage, outperforming $\{\pi_{178}\}$ and $\{\pi_{316}\}$ by $3.9\%$ and $2.4\%$ in validation accuracy, respectively. This collapse under length limits indicates that higher token entropy alone does not translate to better distillation performance. Additionally, we perform an auxiliary experiment where we sample responses for the Easy problem $\pi_{178}$ under higher temperatures to artificially increase token entropy. As shown in Figure \ref{fig:temperature}, the resulting model shows no accuracy gains, indicating that the performance improvement of difficult tasks is not driven by their higher token entropy.

\begin{wrapfigure}{r}{0.4\linewidth} 
    \vspace{-1em}
    \centering
    \includegraphics[width=\linewidth]{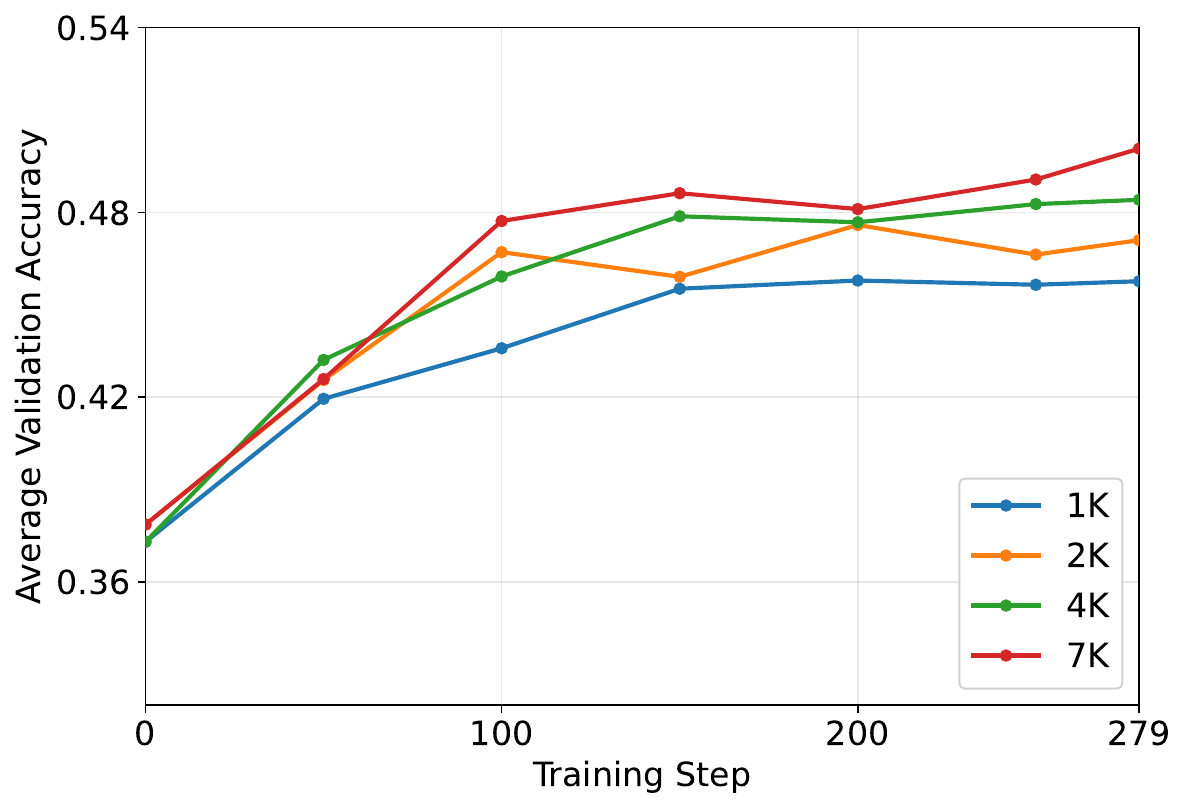}
    \caption{\textbf{Validation performance across different maximum response length.} We compare distillation validation accuracy across training steps under $1\text{K}$, $2\text{K}$, $4\text{K}$, and $7\text{K}$ maximum token limits. }
    \label{fig:response_length_curves}
    \vspace{-1em}
\end{wrapfigure}

\textbf{Distillation Benefits from Longer CoTs} We further compare student models distilled on the hard problem $\{\pi_{954}\}$ under different maximum response length of 1024, 2048, 4096, and 7168 tokens (referred to as 1K, 2K, 4K, and 7K models, respectively). Figure \ref{fig:response_length_curves} illustrates the validation accuracy dynamics across training steps under these length limitation. The empirical results show a clear trend within the 1K to 7K token range: increasing the maximum response length enables the student model to engage in longer reasoning processes, leading to progressively higher validation accuracy. To explore the underlying mechanics, we track the token level KL loss over generation position in Figure \ref{fig:longer_cot_kl_loss}. As depicted in Figure \ref{fig:longer_cot_kl_loss}, the 7K model maintains a consistently lower KL divergence across extended generation horizons, whereas shorter variants exhibit a sharp rise in KL loss as the generation sequence lengthens. This indicates that training on longer CoT sequences directly enhances the model's longer-horizon reasoning capabilities, enabling it to sustain policy alignment with the teacher over extended context lengths.

\textbf{Harder Problems can Trigger More Reasoning Patterns.} Furthermore, we observe that training on harder problems does not simply scale token volume, but instead learns more reasoning patterns, such as self-reflection and backtracking. As shown in Figure \ref{fig:token_kl_reduction}, for the model training on easy problem ($\{\pi_{178}\}$) where trajectories are naturally short, the self-reflection token \texttt{"Alternatively"} is absent from the top-16 KL reduction list. However, it emerges in the medium model ($\{\pi_{794}\}$), and ranks higher in the hard model ($\{\pi_{954}\}$). This finding suggests that longer CoT trajectories allows the student model to acquire some sophisticated reasoning patterns that are inherently absent in shorter, simpler reasoning paths.

\section{How Much Data is Sufficient for OPD?}  \label{sec:analysis}

\subsection{Few-Shot OPD Can Match Full-set Performance}
\label{subsubsec:few_shot_scaling}

Based on Section \ref{sec:selection}, we simply select hard examples as training set and investigate how scaling the training set size affects performance. Specifically, we scale the number of Hard training examples $N \in \{1, 4, 8, 16, 64\}$. To construct our training sets, for $N = 1, 4,$ and $8$, we directly use the hard examples evaluated in Table~\ref{tab:opd_accuracies} (i.e., $\{\pi_{973}\}$ for 1-shot, $\{\pi_{874}, \pi_{890}, \pi_{954}, \pi_{973}\}$ for 4-shot, and $\{\pi_{874}, \dots, \pi_{997}\}$ for 8-shot). For the larger sizes of $N = 16$ and $64$, in addition to 8 hard examples in Table~\ref{tab:opd_accuracies}, we randomly sample the remaining examples from our hard example pool. For comparison, we evaluate Easy and Medium baselines of size $N=8$ using the eight corresponding examples (i.e., $\{ \pi_{5}, \pi_{20},\dots, \pi_{192}\}$ for Easy, and $\{\pi_{316}, \pi_{487}, \dots, \pi_{820}\}$ for Medium). Additionally, we also construct a Random baseline by randomly sampling 8 examples from the hold out training pool. Table~\ref{tab:few_shot_performance} reports these results. To ensure that our data selection strategy is robust against sampling randomness, we further evaluate its stability across three independent random seeds in Appendix \ref{app:random_seed}. To evaluate whether our methodology generalizes beyond mathematical reasoning, we also evaluate data selection in the code generation domain, with complete experimental setups and results presented in Appendix \ref{app:coding}.

From the results in Table~\ref{tab:few_shot_performance}, we find that: 1) increasing the training set of Hard examples from $N=1$ to $N=8$ steadily improves performance, with 8 examples already matching the full-dataset (17K) distillation baseline, while further increasing the size to 16 or 64 examples yields no additional gains; and 2) consistent with our 1-shot findings, under a fixed size of 8 examples, training on Hard examples yields the best performance over the Easy, Medium, and Random baselines.

\begin{table}[htbp]
\centering
\caption{\textbf{Few-Shot OPD Performance across Different Dataset Sizes.} The baselines include the full-dataset distillation (DAPO-Math-17K) and a randomly sampled 1K subset (DAPO-Subset). Distilling on just 8 hard examples achieves $53.6\%$ average accuracy, nearly matching the Full-Set 17K baseline ($53.7\%$).}
\label{tab:few_shot_performance}
\small
\resizebox{\textwidth}{!}{
\begin{tabular}{lcccccccc}
\toprule
\textbf{Dataset} & \textbf{Size} & \textbf{AIME24} & \textbf{AIME25} & \textbf{AMC23} & \textbf{MATH500} & \textbf{Olympiad} & \textbf{Minerva} & \textbf{Avg} \\
\midrule
DAPO-Math-17K & 17K & 51.9 & 34.0 & 78.9 & 86.7 & 44.3 & 26.6 & 53.7  \\
DAPO-Subset & 1K & 50.2 & 34.8 & 79.3 & 87.1 & 42.8 & 27.3 & 53.6 \\
\midrule
$\{\pi_{5}, \pi_{20}, \dots, \pi_{192}\}$ & 8 & 45.8 & 30.6 & 76.1 & 87.1 & 41.6 & 26.2 & 51.2 \\
$\{\pi_{316}, \pi_{487}, \dots, \pi_{820}\}$ & 8 & 49.0 & 32.7 & 76.8 & 86.8 & 42.1 & 26.0 & 52.2 \\
Random & 8 & 48.1 & 33.8 & 76.2 & 86.3 & 41.9 & 26.4 & 52.1 \\ 
\midrule
$\{\pi_{973}\}$  & 1 & 47.5 & 31.7 & 75.1 & 87.7 & 41.5 & 26.9 & 51.7 \\
$\{\pi_{874}, \pi_{890}, \pi_{954}, \pi_{973}\}$ & 4 & 50.8 & 31.5 & 78.5 & 86.7 & 41.5 & 28.2 & 52.9 \\
$\{\pi_{874}, \dots, \pi_{997}\}$ & 8 & 51.7 & 34.2 & 79.2 & 87.2 & 42.7 & 26.4 & 53.6 \\
$\{\pi_{874}, \dots, \pi_{997}, \dots\}$ & 16 & 49.4 & 33.3 & 78.5 & 87.7 & 43.0 & 27.7 & 53.3 \\
$\{\pi_{874}, \dots, \pi_{997}, \dots\}$ & 64 & 50.2 & 36.7 & 79.1 & 87.8 & 42.7 & 27.1 & 53.9 \\
\bottomrule
\end{tabular}}
\end{table}

\subsection{Few-Shot OPD on Other Models}
\label{subsec:other_models}

To investigate whether this extreme data efficiency generalizes across other models, we additionally evaluate four teacher–student pairs: 1) \textbf{Skywork-OR1-Math-7B $\rightarrow$ DeepSeek-R1-Distill-Qwen-7B}; 2) \textbf{DeepSeek-R1-Distill-Llama-8B-GRPO $\rightarrow$ DeepSeek-R1-Distill-Llama-8B}, where the former is obtained by applying GRPO to the latter (detailed training settings are provided in Appendix \ref{app:grpo_training_details}); 3) \textbf{Qwen3-4B $\rightarrow$ Qwen3-1.7B-Base}; 4) \textbf{Qwen3-4B $\rightarrow$ Qwen3-4B-Base}. The setting vary scale, capability, and backbone architecture. For Qwen3-4B, we disable thinking mode. Similar to Section \ref{subsubsec:few_shot_scaling}, we train the student models on either 8 selected hard examples ($\{\pi_{874}, \dots, \pi_{997}\}$) or 8 randomly sampled examples to compare their distillation performance. 
As shown in Table \ref{tab:cross_model_performance}, few shot OPD on selected hard examples consistently triggers substantial performance gains, outperforming the Random baseline and closely approaching full dataset distillation.

\begin{table}[t!]
\centering
\caption{\textbf{Selecting few hard examples for OPD is effective across different models.} We report benchmark accuracies across diverse model pairings and training sizes. }
\label{tab:cross_model_performance}
\small
\resizebox{\textwidth}{!}{
\begin{tabular}{lcccccccc}
\toprule
\textbf{Model} & \textbf{Size} & \textbf{AIME24} & \textbf{AIME25} & \textbf{AMC23} & \textbf{MATH500} & \textbf{Olympiad} & \textbf{Minerva} & \textbf{Avg} \\
\midrule
\midrule
\multicolumn{9}{l}{\textbf{Skywork-OR1-Math-7B $\rightarrow$ DeepSeek-R1-Distill-Qwen-7B}} \\
\midrule
\midrule
DeepSeek-R1-Distill-Qwen-7B & -- & 51.9 & 38.8 & 81.0 & 89.6 & 43.6 & 32.6 & 56.3 \\
Skywork-OR1-Math-7B & -- & 62.1 & 45.6 & 84.0 & 91.9 & 47.7 & 34.1 & 60.9 \\
\midrule
Full-Set & 17K & 58.3 & 44.6 & 83.4 & 91.1 & 46.0 & 34.2 & 59.6 \\
\midrule
Random & 8 & 57.3 & 42.1 & 81.1 & 91.0 & 45.2 & 33.2 & 58.3 \\
$\{\pi_{874}, \dots, \pi_{997}\}$ & 8 & 58.3 & 43.9 & 83.1 & 91.3 & 46.3 & 34.0 & 59.5 \\
\midrule
\midrule
\multicolumn{9}{l}{\textbf{DeepSeek-R1-Distill-Llama-8B-GRPO $\rightarrow$ DeepSeek-R1-Distill-Llama-8B}} \\
\midrule
\midrule
DeepSeek-R1-Distill-Llama-8B & -- & 46.0 & 28.4 & 78.5 & 86.0 & 41.9 & 27.1 & 51.3 \\
DeepSeek-R1-Distill-Llama-8B-GRPO & -- & 50.8 & 32.4 & 85.3 & 87.7 & 44.2 & 28.7 & 54.9 \\
\midrule
Full-Set & 17K & 50.1 & 31.2 & 84.7 & 87.4 & 43.0 & 28.4 & 54.1 \\
\midrule
Random & 8 & 48.9 & 29.7 & 82.2 & 86.7 & 42.3 & 27.7 & 52.9 \\
$\{\pi_{874}, \dots, \pi_{997}\}$ & 8 & 50.3 & 30.9 & 84.3 & 87.1 & 42.8 & 28.1 & 53.9 \\
\midrule
\midrule
\multicolumn{9}{l}{\textbf{Qwen3-4B $\rightarrow$ Qwen3-1.7B-Base}} \\
\midrule
\midrule
Qwen3-1.7B-Base & -- & 1.4 & 2.5 & 12.3 & 13.3 & 4.8 & 3.7 & 6.3 \\
Qwen3-4B & -- & 22.9 & 17.1 & 62.0 & 81.1 & 45.4 & 26.1 & 42.4 \\
\midrule
Full-Set & 17K & 10.2 & 4.0 & 28.6 & 53.9 & 24.4 & 14.1 & 22.5 \\
\midrule
Random & 8 & 6.5 & 3.6 & 24.6 & 46.0 & 22.3 & 11.1 & 19.0 \\
$\{\pi_{874}, \dots, \pi_{997}\}$ & 8 & 9.4 & 4.4 & 26.8 & 51.0 & 23.7 & 12.7 & 21.3 \\
\midrule
\midrule
\multicolumn{9}{l}{\textbf{Qwen3-4B $\rightarrow$ Qwen3-4B-Base}} \\
\midrule
\midrule
Qwen3-4B-Base & -- & 7.3 & 5.0 & 24.6 & 22.3 & 11.9 & 4.8 & 12.7 \\
Qwen3-4B & -- & 22.9 & 17.1 & 62.0 & 81.1 & 45.4 & 26.1 & 42.4 \\
\midrule
Full-Set & 17K & 14.4 & 14.4 & 40.5 & 65.6 & 33.3 & 16.5 & 30.8 \\
\midrule
Random & 8 & 11.9 & 10.6 & 36.1 & 59.3 & 29.1 & 15.0 & 27.0 \\
$\{\pi_{874}, \dots, \pi_{997}\}$ & 8 & 13.1 & 12.5 & 38.6 & 64.1 & 31.3 & 15.6 & 29.2 \\
\bottomrule
\end{tabular}}
\end{table}

\section{Related Work}
\paragraph{On-Policy Distillation.} Distillation for LLMs is broadly categorized into off-policy and on-policy settings. While off-policy distillation (e.g., SFT) suffers from exposure bias due to the mismatch between static teacher-generated data and student-generated rollouts at inference \citep{agarwal2024policy}, OPD mitigates this issue by training on trajectories sampled directly from the student's current policy under a reverse KL objective \citep{gu2024minillm}. Recently, OPD has since appeared in some reasoning and post-training recipes, such as Qwen3 \citep{yang2025qwen3}, Deepseek-V4 \citep{xu2026deepseek} and GLM-5 \citep{zeng2026glm}. Building on these successes, a growing body of work has investigated various algorithmic variants of OPD \citep{oh2026kl, yang2026prune}, optimized its practical training recipes \citep{li2026filter, ko2026scaling}, and integrated it into large-scale post-training pipelines \citep{ma2026mopd}. Moreover, several studies have further analyzed the underlying mechanisms of OPD \citep{wang2026not, li2026rethinking}. 

\paragraph{Data Selection for LLM Post-Training.} Selecting a small yet effective subset of training data has become an important problem for improving the efficiency of LLM post-training, with most efforts focusing on data selection for SFT \citep{ivison2025large} and RLVR \citep{wu2026single}. Existing approaches include LLM-based quality assessment \citep{chen2024alpagasus}, gradient-based selection \citep{xia2024less} and more. However, data selection for OPD remains underexplored. Our work is inspired by recent advances in RLVR, which show that RLVR can achieve significant reasoning gains with as few as a single training example \citep{wang2026reinforcement}. We extend this exploration to the OPD paradigm, discovering different and interesting phenomena. A concurrent work by \citet{fu2026rethinking} also investigates 1-shot OPD and analyzes its efficacy from a state-space coverage perspective. Differently, we analyze 1-shot OPD from the perspective of CoT trajectory and reasoning patterns. Then we propose a simple and effective data selection method that selects only hard examples for training, where even ``unsolvable'' examples completely exceeding the teacher's capability can be successfully used.

\section{Conclusions} 
In this work, we show that 1-shot OPD is sufficient to trigger substantial improvements in mathematical reasoning tasks, with as few as 8 selected examples nearly matching the performance of distilling on the full DAPO-Math-17K dataset. Moreover, we observe that distilling on harder tasks consistently outperforms easy ones, and even ``unsolvable'' tasks where both the teacher and student models fail to produce correct answers still yield strong performance gains. Our analysis reveals that this improvement is fundamentally driven by the longer CoT trajectories rather than high token-level entropy. Future work includes: 1) designing effective dynamic batching strategies algorithms of OPD based on our findings; and 2) developing novel OPD algorithms that can learn from longer CoT reasoning trajectories, where the teacher’s dense reward may lose local exploitability.

\bibliography{opd}

@article{li2026rethinking,
  title={Rethinking on-policy distillation of large language models: Phenomenology, mechanism, and recipe},
  author={Li, Yaxuan and Zuo, Yuxin and He, Bingxiang and Zhang, Jinqian and Xiao, Chaojun and Qian, Cheng and Yu, Tianyu and Gao, Huan-ang and Yang, Wenkai and Liu, Zhiyuan and others},
  journal={arXiv preprint arXiv:2604.13016},
  year={2026}
}

@article{yu2026dapo,
  title={Dapo: An open-source llm reinforcement learning system at scale},
  author={Yu, Qiying and Zhang, Zheng and Zhu, Ruofei and Yuan, Yufeng and Zuo, Xiaochen and Yue, Yu and Dai, Weinan and Fan, Tiantian and Liu, Gaohong and Liu, Lingjun and others},
  journal={Advances in Neural Information Processing Systems},
  volume={38},
  pages={113222--113244},
  year={2026}
}

@article{sheng2024hybridflow,
  title   = {HybridFlow: A Flexible and Efficient RLHF Framework},
  author  = {Guangming Sheng and Chi Zhang and Zilingfeng Ye and Xibin Wu and Wang Zhang and Ru Zhang and Yanghua Peng and Haibin Lin and Chuan Wu},
  year    = {2024},
  journal = {arXiv preprint arXiv: 2409.19256}
}

@inproceedings{hendrycks2measuring,
  title={Measuring Mathematical Problem Solving With the MATH Dataset},
  author={Hendrycks, Dan and Burns, Collin and Kadavath, Saurav and Arora, Akul and Basart, Steven and Tang, Eric and Song, Dawn and Steinhardt, Jacob},
  year = {2021},
  booktitle={Thirty-fifth Conference on Neural Information Processing Systems Datasets and Benchmarks Track (Round 2)}
}

@misc{aime,
  author       = {{Art of Problem Solving}},
  title        = {AIME Problems and Solutions},
  howpublished = {\url{https://artofproblemsolving.com/wiki/index.php/AIME_Problems_and_Solutions}},
  note         = {Accessed: 2025-04-20}
}

@misc{amc,
  author       = {{Art of Problem Solving}},
  title        = {AMC Problems and Solutions},
  howpublished = {\url{https://artofproblemsolving.com/wiki/index.php?title=AMC_Problems_and_Solutions}},
  note         = {Accessed: 2025-04-20}
}

@article{lewkowycz2022solving,
  title={Solving quantitative reasoning problems with language models},
  author={Lewkowycz, Aitor and Andreassen, Anders and Dohan, David and Dyer, Ethan and Michalewski, Henryk and Ramasesh, Vinay and Slone, Ambrose and Anil, Cem and Schlag, Imanol and Gutman-Solo, Theo and others},
  journal={Advances in neural information processing systems},
  volume={35},
  pages={3843--3857},
  year={2022}
}

@inproceedings{he2024olympiadbench,
  title={Olympiadbench: A challenging benchmark for promoting agi with olympiad-level bilingual multimodal scientific problems},
  author={He, Chaoqun and Luo, Renjie and Bai, Yuzhuo and Hu, Shengding and Thai, Zhen and Shen, Junhao and Hu, Jinyi and Han, Xu and Huang, Yujie and Zhang, Yuxiang and others},
  booktitle={Proceedings of the 62nd Annual Meeting of the Association for Computational Linguistics (Volume 1: Long Papers)},
  pages={3828--3850},
  year={2024}
}

@inproceedings{jainlivecodebench,
  title={LiveCodeBench: Holistic and Contamination Free Evaluation of Large Language Models for Code},
  author={Jain, Naman and Han, King and Gu, Alex and Li, Wen-Ding and Yan, Fanjia and Zhang, Tianjun and Wang, Sida and Solar-Lezama, Armando and Sen, Koushik and Stoica, Ion},
  year = {2025},
  booktitle={The Thirteenth International Conference on Learning Representations}
}

@article{wang2026reinforcement,
  title={Reinforcement learning for reasoning in large language models with one training example},
  author={Wang, Yiping and Yang, Qing and Zeng, Zhiyuan and Ren, Liliang and Liu, Liyuan and Peng, Baolin and Cheng, Hao and He, Xuehai and Wang, Kuan and Gao, Jianfeng and others},
  journal={Advances in Neural Information Processing Systems},
  volume={38},
  pages={122721--122764},
  year={2026}
}

@article{jin2026entropy,
  title={Entropy-aware on-policy distillation of language models},
  author={Jin, Woogyeol and Min, Taywon and Yang, Yongjin and Wei, Dennis and Zhou, Yi and Kadhe, Swanand Ravindra and Baracaldo, Nathalie and Lee, Kimin},
  journal={arXiv preprint arXiv:2603.07079},
  year={2026}
}

@article{ko2026scaling,
  title={Scaling reasoning efficiently via relaxed on-policy distillation},
  author={Ko, Jongwoo and Abdali, Sara and Kim, Young Jin and Chen, Tianyi and Cameron, Pashmina},
  journal={arXiv preprint arXiv:2603.11137},
  year={2026}
}

@article{yang2025qwen3,
  title={Qwen3 technical report},
  author={Yang, An and Li, Anfeng and Yang, Baosong and Zhang, Beichen and Hui, Binyuan and Zheng, Bo and Yu, Bowen and Gao, Chang and Huang, Chengen and Lv, Chenxu and others},
  journal={arXiv preprint arXiv:2505.09388},
  year={2025}
}

@article{xiao2026mimo,
  title={Mimo-v2-flash technical report},
  author={Xiao, Bangjun and Xia, Bingquan and Yang, Bo and Gao, Bofei and Shen, Bowen and Zhang, Chen and He, Chenhong and Lou, Chiheng and Luo, Fuli and Wang, Gang and others},
  journal={arXiv preprint arXiv:2601.02780},
  year={2026}
}

@article{zeng2026glm,
  title={Glm-5: from vibe coding to agentic engineering},
  author={Zeng, Aohan and Lv, Xin and Hou, Zhenyu and Du, Zhengxiao and Zheng, Qinkai and Chen, Bin and Yin, Da and Ge, Chendi and Huang, Chenghua and Xie, Chengxing and others},
  journal={arXiv preprint arXiv:2602.15763},
  year={2026}
}

@article{lu2025onpolicydistillation,
  author = {Kevin Lu and Thinking Machines Lab},
  title = {On-Policy Distillation},
  journal = {Thinking Machines Lab: Connectionism},
  year = {2025},
  note = {https://thinkingmachines.ai/blog/on-policy-distillation},
  doi = {10.64434/tml.20251026},
}

@article{he2025justrl,
  title={Justrl: Scaling a 1.5 b llm with a simple rl recipe},
  author={He, Bingxiang and Qu, Zekai and Liu, Zeyuan and Chen, Yinghao and Zuo, Yuxin and Qian, Cheng and Zhang, Kaiyan and Chen, Weize and Xiao, Chaojun and Cui, Ganqu and others},
  journal={arXiv preprint arXiv:2512.16649},
  year={2025}
}

@article{guo2025deepseek,
  title={Deepseek-r1: Incentivizing reasoning capability in llms via reinforcement learning},
  author={Guo, Daya and Yang, Dejian and Zhang, Haowei and Song, Junxiao and Wang, Peiyi and Zhu, Qihao and Xu, Runxin and Zhang, Ruoyu and Ma, Shirong and Bi, Xiao and others},
  journal={arXiv preprint arXiv:2501.12948},
  year={2025}
}

@inproceedings{agarwal2024policy,
  title={On-policy distillation of language models: Learning from self-generated mistakes},
  author={Agarwal, Rishabh and Vieillard, Nino and Zhou, Yongchao and Stanczyk, Piotr and Ramos Garea, Sabela and Geist, Matthieu and Bachem, Olivier},
  booktitle={International Conference on Learning Representations},
  volume={2024},
  pages={21246--21263},
  year={2024}
}

@inproceedings{gu2024minillm,
  title={Minillm: Knowledge distillation of large language models},
  author={Gu, Yuxian and Dong, Li and Wei, Furu and Huang, Minlie},
  booktitle={International Conference on Learning Representations},
  volume={2024},
  pages={32694--32717},
  year={2024}
}

@article{xu2026deepseek,
  title={Deepseek-v4: Towards highly efficient million-token context intelligence},
  author={Xu, Anyi and Lin, Bangcai and Xue, Bing and Wang, Bingxuan and Xu, Bingzheng and Wu, Bochao and Zhang, Bowei and Lin, Chaofan and Dong, Chen and Ling, Chenchen and others},
  journal={arXiv preprint arXiv:2606.19348},
  year={2026}
}

@article{ma2026mopd,
  title={Mopd: Multi-teacher on-policy distillation for capability integration in llm post-training},
  author={Ma, Wenhan and Wei, Jianyu and Zhao, Liang and Zhang, Hailin and Xiao, Bangjun and Li, Lei and Yang, Qibin and Gao, Bofei and Wang, Yudong and Li, Rang and others},
  journal={arXiv preprint arXiv:2606.30406},
  year={2026}
}

@article{oh2026kl,
  title={KL for a KL: On-Policy Distillation with Control Variate Baseline},
  author={Oh, Minjae and Song, Sangjun and Choi, Gyubin and Choi, Yunho and Jo, Yohan},
  journal={arXiv preprint arXiv:2605.07865},
  year={2026}
}

@article{yang2026prune,
  title={Prune-OPD: Efficient and Reliable On-Policy Distillation for Long-Horizon Reasoning},
  author={Yang, Zhicheng and Guo, Zhijiang and Song, Yifan and Xu, Minrui and Wang, Yongxin and Wang, Yiwei and Liang, Xiaodan and Tang, Jing},
  journal={arXiv preprint arXiv:2605.07804},
  year={2026}
}

@article{li2026filter,
  title={Filter, then reweight: Rethinking optimization granularity in on-policy distillation},
  author={Li, Yuying and Zheng, Leqi and Yu, Yongzi and Zhou, Wenrui and Zhong, Xuchang and Hu, Xing and Jin, Jing and Yuan, Hangjie and Feng, Tao},
  journal={arXiv preprint arXiv:2606.02684},
  year={2026}
}

@article{wang2026not,
  title={Not all disagreement is learnable: Token teachability in on-policy distillation},
  author={Wang, Yuanyi and Lu, Su and Gu, Yanggan and Wang, Pengkai and Yang, Yifan and Yan, Zhaoyi and Xie, Congkai and Wu, Jianmin and Yang, Hongxia},
  journal={arXiv preprint arXiv:2605.26844},
  year={2026}
}

@article{wu2026single,
  title={Single-Rollout Hidden-State Dynamics for Training-Free RLVR Data Selection},
  author={Wu, Jianghao and Cai, Jianfei and Wang, Weiqiang and Ye, Jin and Schmidt, Daniel F and George, Yasmeen},
  journal={arXiv preprint arXiv:2605.28631},
  year={2026}
}

@article{ivison2025large,
  title={Large-scale data selection for instruction tuning},
  author={Ivison, Hamish and Zhang, Muru and Brahman, Faeze and Koh, Pang Wei and Dasigi, Pradeep},
  journal={arXiv preprint arXiv:2503.01807},
  year={2025}
}

@inproceedings{chen2024alpagasus,
  title={Alpagasus: Training a better alpaca with fewer data},
  author={Chen, Lichang and Li, Shiyang and Yan, Jun and Wang, Hai and Gunaratna, Kalpa and Yadav, Vikas and Tang, Zheng and Srinivasan, Vijay and Zhou, Tianyi and Huang, Heng and others},
  booktitle={International Conference on Learning Representations},
  volume={2024},
  pages={34767--34797},
  year={2024}
}

@article{xia2024less,
  title={Less: Selecting influential data for targeted instruction tuning},
  author={Xia, Mengzhou and Malladi, Sadhika and Gururangan, Suchin and Arora, Sanjeev and Chen, Danqi},
  journal={arXiv preprint arXiv:2402.04333},
  year={2024}
}

@article{fu2026rethinking,
  title={Rethinking on-policy distillation of large language models ii: One training example},
  author={Fu, Zixuan and He, Bingxiang and Zuo, Yuxin and Huang, Haohuan and Zhang, Jinqian and Xiao, Ruhang and Qian, Cheng and Luo, Qinyu and Gao, Huan-ang and Wang, Yudong and others},
  journal={arXiv preprint arXiv:2609.04172},
  year={2026}
}
\bibliographystyle{iclr2027_conference}

\appendix
\newpage
\section*{Appendix Table of Contents}
\startcontents[appendix]
\printcontents[appendix]{l}{1}{\setcounter{tocdepth}{2}}
\newpage

\section{More Training Details} \label{app:training_details}

\subsection{OPD Training Details} \label{app:opd_training_details}
For every input prompt during the OPD training, we generate $n=8$ responses. We constrain the maximum prompt length to 1,024 tokens, while the maximum response length is limited to 7,168 tokens. The optimization process spans a single epoch across 8 H800 80G GPUs, utilizing a learning rate of $1 \times 10^{-6}$. We set both the student sampling temperature and the teacher temperature to 1.0, disable KL regularization, and adopt token-mean loss aggregation. Unless otherwise noted, all experiments use the default OPD hyperparameters listed in Table \ref{tab:opd_hyperparameters}.

\begin{table}[htbp]
\centering
\caption{Default hyperparameters for OPD.}
\label{tab:opd_hyperparameters}
\begin{tabular}{lc}
\toprule
\textbf{Item} & \textbf{Value} \\
\midrule
Training temperature & 1.0 \\
Global batch size & 64 \\
Mini batch size & 64 \\
Rollout number & 8 \\
Temperature & 1.0 \\
LogProb top-$K$ & 16 \\
Top-$K$ strategy & Student Top-$K$ \\
Top-$p$ & 1.0 \\
Max prompt length & 1024 \\
Max response length & 7168 \\
Learning rate & 1e-6 \\
Epoch & 1 \\
KL Coefficient & 0.0 \\
\bottomrule
\end{tabular}
\end{table}

\subsection{GRPO Training Details} \label{app:grpo_training_details}
We train DeepSeek-R1-Distill-Llama-8B for one epoch on the dataset 
DeepMath-103K \citep{deepmath}. The main hyperparameters are summarized in Table \ref{tab:grpo_hyperparameters}.

\begin{table}[htbp]
\centering
\caption{Default hyperparameters for GRPO.}
\label{tab:grpo_hyperparameters}
\begin{tabular}{lc}
\toprule
\textbf{Item} & \textbf{Value} \\
\midrule
Training temperature & 1.0 \\
Global batch size & 64 \\
Mini batch size & 64 \\
Rollout number & 8 \\
Temperature & 1.0 \\
Max prompt length & 1024 \\
Max response length & 7168 \\
Learning rate & 1e-6 \\
Epoch & 1 \\
KL Coefficient & 0.0 \\
\bottomrule
\end{tabular}
\end{table}

\section{More Evaluations} \label{app:ood}
To verify whether the performance gains acquired through 1-shot OPD generalize beyond the mathematical domain, we evaluate our models on out-of-distribution benchmarks, including LiveCodeBench for code reasoning and GPQA for scientific reasoning. Specifically, we report the LiveCodeBench and GPQA accuracy for 1-shot OPD models from Table \ref{tab:one_shot_benchmark_performance} in Table \ref{tab:ood_1shot_models}, and the performance of few-shot OPD models from Table \ref{tab:few_shot_performance} in Table \ref{tab:ood_fewshot_models}. The empirical results further validate that our observations remain highly consistent across out-of-distribution tasks. First, 1-shot OPD trained on a single mathematical example successfully transfers reasoning abilities to coding and scientific domain, yielding notable performance gains over the base student model. As shown in Table \ref{tab:ood_1shot_models}, the 1-shot OPD training on single hard problems achieves an average accuracy of $40.5\%$ on LiveCodeBench and $24.2\%$ on GPQA, significantly outperforming the base student model baselines of $32.4\%$ and $19.4\%$, respectively. Second, training on harder problems consistently achieves superior cross domain performance compared to easy problems. According to Table \ref{tab:ood_1shot_models}, the average accuracy of 1-shot OPD on hard problems ($40.5\%$ on LiveCodeBench and $24.2\%$ on GPQA) markedly surpasses that training on easy problems ($37.6\%$ on LiveCodeBench and $22.1\%$ on GPQA). Third, distilling on 8 selected hard examples is sufficient to closely approach the full dataset baseline. As reported in Table \ref{tab:ood_fewshot_models}, it reaches $42.0\%$ on LiveCodeBench and $27.2\%$ on GPQA, nearly matching the full 17K dataset performance of $42.2\%$ on LiveCodeBench and $27.8\%$ on GPQA.

\begin{table}[htbp]
\centering
\caption{\textbf{Out of Distribution Generalization of 1-Shot OPD Models from Table \ref{tab:one_shot_benchmark_performance}.} We report the accuracy on LiveCodeBench (LCB) and GPQA across individual models. The \textbf{Avg} rows denote the average accuracy across the 8 models trained on Easy, Medium, and Hard samples, respectively.}
\label{tab:ood_1shot_models}
\small
\begin{tabular}{lcccccccr}
\toprule
\multicolumn{3}{c}{\textbf{Easy Problems}} & \multicolumn{3}{c}{\textbf{Medium Problems}} & \multicolumn{3}{c}{\textbf{Hard Problems}} \\
\cmidrule(lr){1-3} \cmidrule(lr){4-6} \cmidrule(lr){7-9}
\textbf{Model} & \textbf{LCB} & \textbf{GPQA} & \textbf{Model} & \textbf{LCB} & \textbf{GPQA} & \textbf{Model} & \textbf{LCB} & \textbf{GPQA} \\
\midrule
$\{\pi_{5}\}$ & 37.3 & 21.4 & $\{\pi_{316}\}$ & 39.3 & 22.2 & $\{\pi_{874}\}$ & 40.9 & 24.6 \\
$\{\pi_{20}\}$ & 38.3 & 21.1 & $\{\pi_{487}\}$ & 39.6 & 23.7 & $\{\pi_{890}\}$ & 40.1 & 23.4 \\
$\{\pi_{21}\}$ & 37.1 & 23.5 & $\{\pi_{530}\}$ & 39.4 & 23.0 & $\{\pi_{948}\}$ & 41.8 & 25.9 \\
$\{\pi_{63}\}$ & 38.8 & 21.6 & $\{\pi_{543}\}$ & 38.9 & 22.3 & $\{\pi_{954}\}$ & 40.8 & 23.7 \\
$\{\pi_{84}\}$ & 36.9 & 22.7 & $\{\pi_{661}\}$ & 39.1 & 23.6 & $\{\pi_{961}\}$ & 39.7 & 23.8 \\
$\{\pi_{105}\}$ & 36.8 & 21.5 & $\{\pi_{763}\}$ & 40.0 & 24.8 & $\{\pi_{973}\}$ & 39.3 & 23.1 \\
$\{\pi_{178}\}$ & 38.7 & 23.1 & $\{\pi_{794}\}$ & 39.4 & 22.7 & $\{\pi_{987}\}$ & 40.4 & 24.3 \\
$\{\pi_{192}\}$ & 37.0 & 21.9 & $\{\pi_{820}\}$ & 38.4 & 22.3 & $\{\pi_{997}\}$ & 40.8 & 24.5 \\
\midrule
\textbf{Avg} & \textbf{37.6} & \textbf{22.1} & \textbf{Avg} & \textbf{39.3} & \textbf{23.1} & \textbf{Avg} & \textbf{40.5} & \textbf{24.2} \\
\bottomrule
\end{tabular}

\caption{\textbf{Out of Distribution Generalization of Few-Shot OPD Models from Table \ref{tab:few_shot_performance}.} We report the average accuracy in LiveCodeBench and GPQA.}
\label{tab:ood_fewshot_models}
\small
\begin{tabular}{lccc}
\toprule
\textbf{Model} & \textbf{Size} & \textbf{LiveCodeBench } & \textbf{GPQA} \\
\midrule
Student & \text{N/A} & 32.4 & 19.4 \\
Teacher & \text{N/A} & 42.8 & 28.0 \\
DAPO-Math-17K & 17K & 42.2 & 27.8 \\
\midrule
$\{\pi_{5}, \pi_{20}, \dots, \pi_{192}\}$ & 8 & 40.6 & 23.7 \\
$\{\pi_{316}, \pi_{487}, \dots, \pi_{820}\}$ & 8 & 40.1 & 24.1 \\
Random & 8 & 40.3 & 24.3 \\
\midrule
$\{\pi_{874}, \pi_{890}, \pi_{954}, \pi_{973}\}$ & 4 & 41.4 & 24.8 \\
$\{\pi_{874}, \dots, \pi_{997}\}$ & 8 & 42.0 & 27.2 \\
$\{\pi_{874}, \dots, \pi_{997}, \dots\}$ & 16 & 40.6 & 26.7 \\
$\{\pi_{874}, \dots, \pi_{997}, \dots\}$ & 64 & 41.7 & 27.5 \\
\bottomrule
\end{tabular}
\end{table}

\section{More Experiments}
\subsection{1-shot OPD in Coding Domain} \label{app:coding}
To evaluate whether our findings generalize beyond mathematical reasoning, we extend our empirical investigation to the code generation domain using the LeetCodeDataset v3.0.1 \citep{xia2025leetcodedataset}, comprising 2,641 problems in total. The dataset provides pre-existing difficulty classifications, containing 638 Easy, 1,397 Medium, and 606 Hard problems. Directly utilizing these native difficulty tags, we sample single training prompts from the Easy, Medium, and Hard categories, respectively. We employ DeepSeek-R1-Distill-Qwen-1.5B as the student model and JustRL-DeepSeek-1.5B as the teacher model. For evaluation, we track LiveCodeBench as the in distribution validation set. To monitor out of distribution performance, we average the accuracy across AMC23, AIME24, and AIME25 as Math Accuracy metric. As illustrated in Figure 6, 1-shot OPD trained on code generation tasks yields substantial validation gains across training steps, while the steady convergence of distillation loss shows effective policy alignment between the student and teacher models. Furthermore, consistent with our mathematical domain observations, distilling on hard coding problems achieves superior performance on both in-distribution and out-of-distribution metrics compared to easy or medium problems.

\begin{figure}[htbp]
    \centering
    \includegraphics[width=\textwidth]{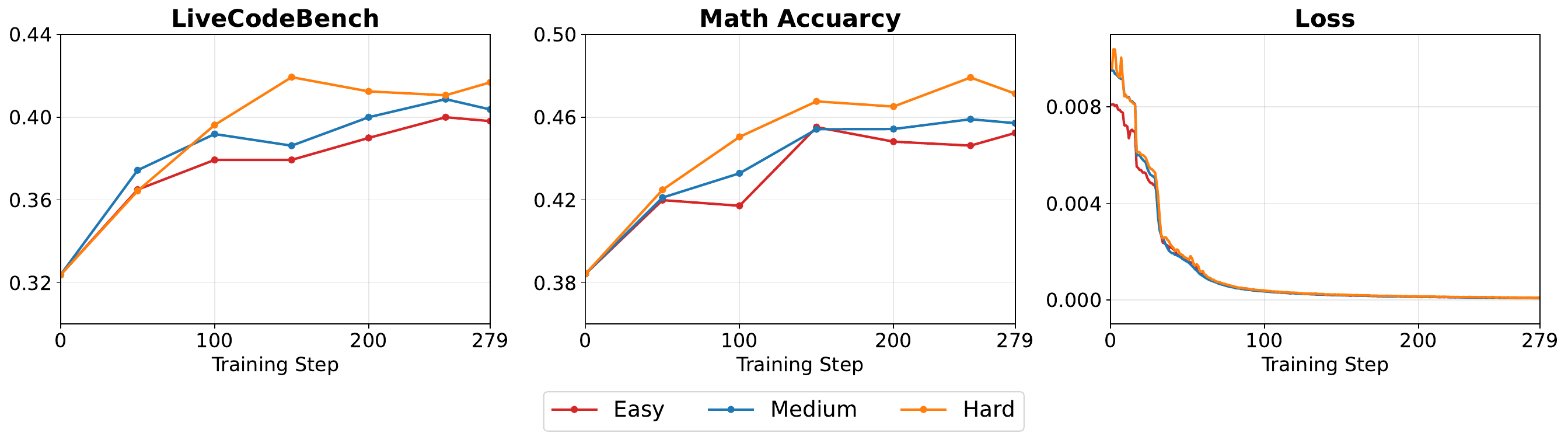}
    \caption{\textbf{Training dynamics and validation trajectories of 1-shot OPD in the coding domain.} We report the in-distribution validation accuracy on LiveCodeBench, the out of distribution Math Accuracy (computed as the average accuracy across AMC23, AIME24, and AIME25), and the distillation loss over training steps. }
    \label{fig:code_training}
\end{figure}

To further evaluate whether our difficulty-driven data selection strategy remains effective in the coding domain, we conduct experiments scaling training prompts from the LeetCodeDataset. Specifically, we randomly sample 8 hard problems to construct our Hard dataset. For comparison, we establish a Full-Set baseline using the entire LeetCodeDataset of 2,641 problems, as well as a Random baseline trained on 8 randomly sampled problems across all difficulty levels. As reported in Table \ref{tab:coding_data_selection}, our difficulty-driven data selection demonstrates efficiency in the coding domain. Training on only 8 selected hard coding problems yields an average accuracy of $47.1\%$, significantly outperforming the Random baseline ($45.1\%$) and nearly recovering the entire performance ceiling of the Full-Set baseline ($47.3\%$). These results indicate that our data selection strategy generalizes robustly to code generation.

\begin{table}[htbp]
\centering
\caption{\textbf{Data Selection Efficacy in the Coding Domain.} We report benchmark performance across six mathematical reasoning benchmarks alongside LiveCodeBench and GPQA. We evaluate the student model (DeepSeek-R1-Distill-Qwen-1.5B) distilled on LeetCodeDataset training prompts under different selection strategies.}
\label{tab:coding_data_selection}
\small
\resizebox{\textwidth}{!}{
\begin{tabular}{lccccccccccc}
\toprule
\textbf{Model} & \textbf{Size} & \textbf{AIME24} & \textbf{AIME25} & \textbf{AMC23} & \textbf{MATH500} & \textbf{Olympiad} & \textbf{Minerva} & \textbf{LiveCodeBench} & \textbf{GPQA} & \textbf{Avg} \\
\midrule
Student & \text{N/A} & 31.7 & 23.0 & 60.7 & 82.6 & 34.1 & 22.4 & 32.4 & 19.4 & 38.3 \\
Teacher & \text{N/A} & 55.8 & 35.8 & 83.4 & 87.4 & 43.4 & 29.8 & 42.8 & 28.0 & 50.8 \\
\midrule
Full-Set & 2641 & 44.4 & 32.1 & 78.7 & 86.3 & 40.8 & 27.0 & 42.7 & 26.4 & 47.3 \\
\midrule
Random & 8 & 41.2 & 29.0 & 72.0 & 86.1 & 39.3 & 26.1 & 41.6 & 25.6 & 45.1 \\
Hard & 8 & \textbf{43.7} & \textbf{31.8} & \textbf{77.2} & \textbf{87.7} & \textbf{39.9} & \textbf{27.0} & \textbf{42.8} & \textbf{26.3} & \textbf{47.1} \\
\bottomrule
\end{tabular}}
\end{table}

\subsection{1-shot OPD over Extended Training}

\begin{figure}[htbp]
    \centering
    \includegraphics[width=\textwidth]{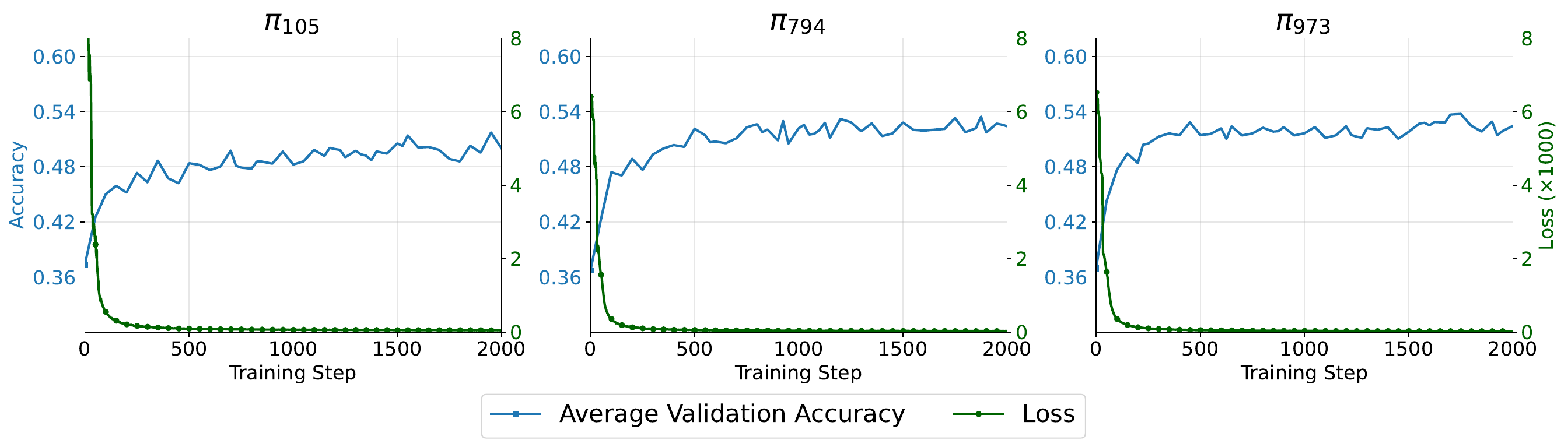}
    \caption{\textbf{Validation Accuracy over Extended Training in 1-Shot OPD.} We plot the validation trajectories over 2,000 steps for 1-shot OPD models trained on three representative examples of varying difficulty levels: $\{\pi_{105}\}$ (Easy, left), $\{\pi_{794}\}$ (Medium, middle), and $\{\pi_{973}\}$ (Hard, right). Despite extended training, the validation performance across all models remains exceptionally stable without policy collapse.}
    \label{fig:long_training}
\end{figure}

We further investigate whether 1-shot OPD suffers from overfitting. To do so, we perform a prolonged training experiment by extending the optimization duration to 2,000 steps for 1-shot OPD training on $\{\pi_{105}\}$, $\{\pi_{794}\}$, and $\{\pi_{973}\}$, respectively. As shown in Figure \ref{fig:long_training}, we observe that the validation accuracy across all three models rises steadily during the first 300 steps and subsequently remains remarkably stable up to 2,000 steps, showing no signs of performance decay or catastrophic policy collapse. This behavior is fundamentally different from SFT and RL in single-example settings \citep{wang2026reinforcement}, which often suffer from severe overfitting or policy degradation under extended optimization. One potential reason is that the sequence-level KL divergence between the teacher and student models naturally decreases as training progresses, leading to loss convergence and preventing policy collapse even under prolonged training.

\subsection{1-shot OPD under Different CoT Lengths} \label{app:kl_position}

\begin{wrapfigure}{r}{0.4\linewidth}
    \vspace{-1em}
    \centering
    \includegraphics[width=\linewidth]{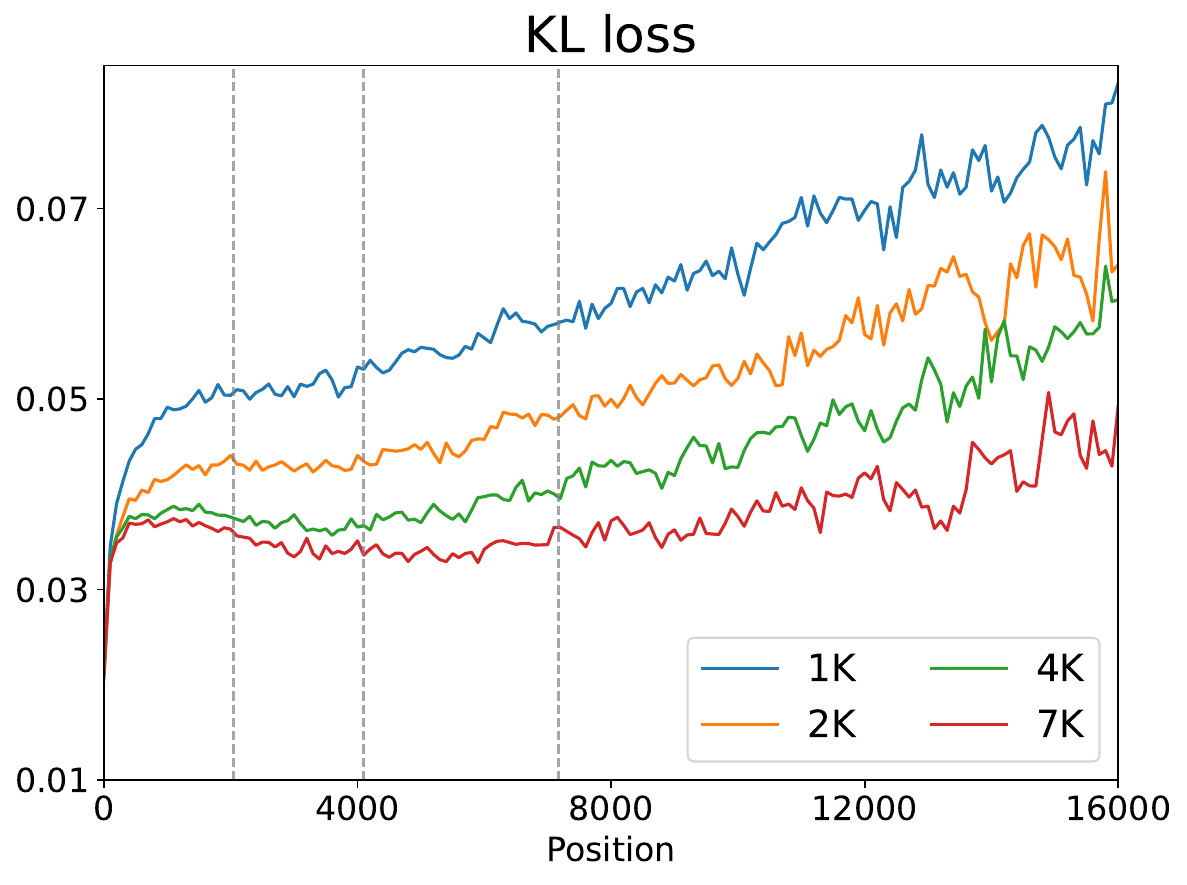}
    \caption{\textbf{Token-level KL Loss over Generation Position.} For visual clarity, we partition the token positions into non-overlapping intervals of 100 tokens and plot the binned average KL loss over positions.}
    \label{fig:longer_cot_kl_loss}
    \vspace{-1em}
\end{wrapfigure}

 We further compare student models distilled on the hard problem $\{\pi_{954}\}$ under maximum rollout length constraints of 1024, 2048, 4096, and 7168 tokens (referred to as the 1K, 2K, 4K, and 7K models, respectively). For evaluation, we generate 16 rollout completions per problem across the 100 held-out problems. At each generation step, we compute the per-token KL divergence between the distilled student and the teacher policies and plot the average KL loss over positions in Figure \ref{fig:longer_cot_kl_loss}.  The empirical results in Figure \ref{fig:longer_cot_kl_loss} reveal that the 7K model consistently maintains a lower KL divergence across the entire generation horizon compared to the 1K, 2K and 4K models. Crucially, while the alignment of the 4K model is nearly identical to that of the 7K model within the $0$ to $4,000$ token range, the KL gap between them widens significantly as the generation horizon expands beyond $4,000$ tokens. This divergence demonstrates that training on longer CoT sequences directly enhances the model's longer-horizon reasoning capabilities, enabling it to sustain policy alignment with the teacher over extended context lengths.

\subsection{1-shot OPD under Different Temperature} 
\begin{figure}[h]
    \centering
    \includegraphics[width=\linewidth]{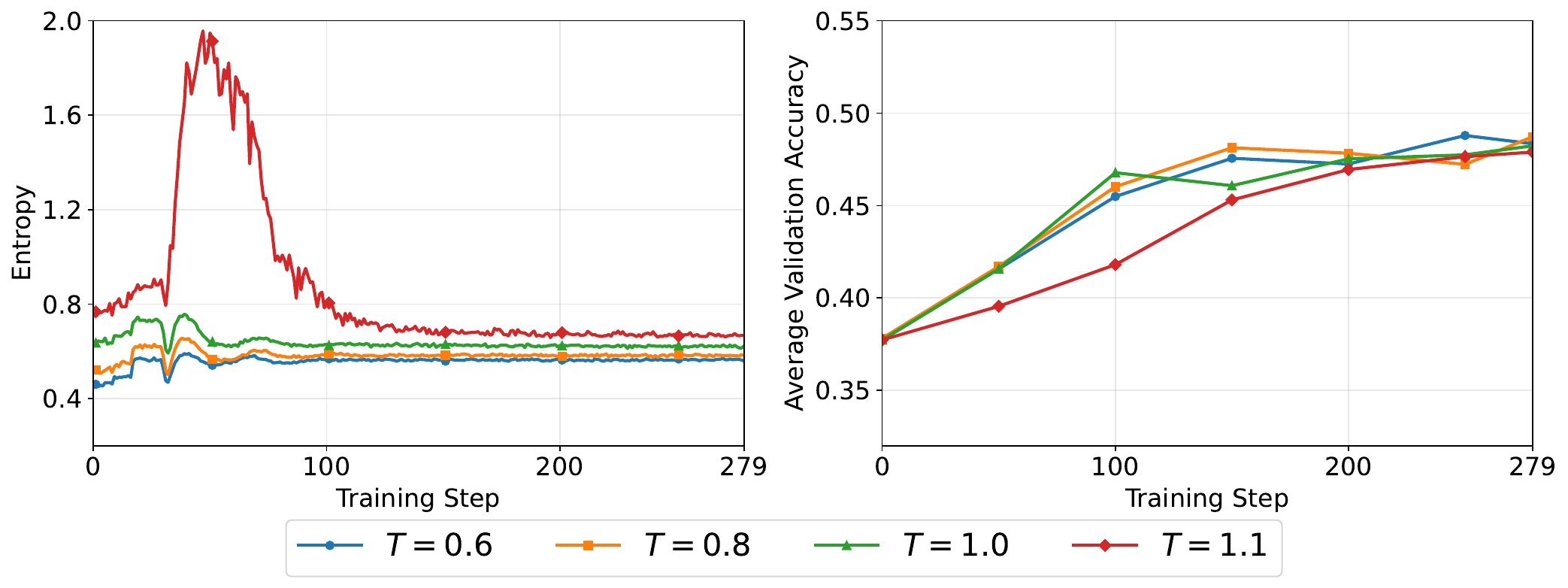}
    \caption{\textbf{Study on student rollout temperature $T$ for the easy example $\pi_{178}$ during 1-shot OPD.} The left subfigure monitors the average per-token entropy of the student's rollouts across training steps under different sampling temperatures $T \in \{0.6, 0.8, 1.0, 1.1\}$. The right subfigure tracks the corresponding Average Validation Accuracy (computed over AMC 2023, AIME 2024, and AIME 2025).}
    \label{fig:temperature}
\end{figure}

We conduct an auxiliary experiment by varying the student rollout temperature $T$ in $\{0.6, 0.8, 1.0, 1.1\}$ during 1-shot OPD. This experiment is specifically designed around one easy problem $\pi_{178}$. By adjusting the sampling temperature of the rollout, we artificially inflate the token-level entropy of the student model's rollouts without altering the inherent difficulty of the training example itself. As illustrated in Figure \ref{fig:temperature}, while higher temperatures markedly elevate the average token entropy throughout training, the corresponding validation accuracy almost keep unchanged across all temperature settings. This empirical result demonstrates that increasing token entropy does not translate to validation gains.

\section{Robustness of Data Selection to Random Seeds} \label{app:random_seed}

To evaluate the stability and robustness of our data selection method against sampling randomness, we extend the setup in Section \ref{subsubsec:few_shot_scaling} by introducing two additional independent random seeds, totaling three independent seeds. Specifically, for each seed, we sample a distinct subset of 8 Hard problems (Hard 8) as well as 8 Random problems (Random 8) from the training pool, where the first seed corresponds to that reported in the main text. All models are evaluated using the student model (DeepSeek-R1-Distill-Qwen-1.5B) and the teacher model (JustRL-DeepSeek-1.5B) across eight benchmarks, comprising six mathematical reasoning tasks alongside LiveCodeBench and GPQA. As reported in Table \ref{tab:seed_robustness}, the performance of our data selection consistently outperforms the random selection across all seeds.

\begin{table}[htbp]
\centering
\caption{\textbf{Robustness of Data Selection Across Different Random Seeds.} We report benchmark accuracies across eight evaluation benchmarks alongside their average (Avg) for different data selection under three independent random sampling seeds. Group 1 represents the default seed used in the main text, while Group 2 and Group 3 represent additional independent random seeds.}
\label{tab:seed_robustness}
\small
\resizebox{\textwidth}{!}{
\begin{tabular}{lcccccccccc}
\toprule
\textbf{Model} & \textbf{AIME24} & \textbf{AIME25} & \textbf{AMC23} & \textbf{MATH500} & \textbf{Olympiad} & \textbf{Minerva} & \textbf{LiveCodeBench} & \textbf{GPQA} & \textbf{Avg} \\
\midrule
\multicolumn{10}{l}{\textit{Random Data Selection}} \\
Seed 1 & 48.1 & 33.8 & 76.2 & 86.3 & 41.9 & 26.4 & 40.3 & 24.3 & 47.2 \\
Seed 2 & 46.7 & 35.0 & 76.1 & 87.0 & 41.6 & 26.1 & 39.9 & 21.9 & 46.8 \\
Seed 3 & 46.2 & 33.8 & 77.0 & 86.4 & 42.4 & 26.2 & 40.1 & 24.9 & 47.1 \\
\midrule
\multicolumn{10}{l}{\textit{Our Data Selection}} \\
Seed 1 & 51.7 & 34.2 & 79.2 & 87.2 & 42.7 & 26.4 & 42.0 & 27.2 & \textbf{48.8} \\
Seed 2 & 50.3 & 34.3 & 78.8 & 87.1 & 42.0 & 28.1 & 41.7 & 27.6 & \textbf{48.7} \\
Seed 3 & 51.3 & 34.9 & 79.3 & 87.5 & 42.3 & 27.8 & 42.2 & 27.8 & \textbf{49.1} \\
\bottomrule
\end{tabular}}
\end{table}

\section{Limitations}

A key limitation of our work is that our difficulty-driven data selection is a simple empirical choice rather than a mathematically optimal one. Finding the theoretically optimal training sets for OPD may be computationally intractable. However, despite not being theoretically optimal, our method is extremely simple to use. It requires zero complex optimization, zero extra training parameters, and can be implemented simply by sorting accuracy of examples. We believe this extreme simplicity and ease of implementation make our findings highly valuable for practical applications.

\section{Discussions}

\paragraph{CoT Length as a Practical Proxy in Verifier-Free Domains.}
In mathematical and coding domains, rule-based verifiers are readily available to compute rollout accuracy for task selection. However, in many general reasoning domains, such verifiers do not exist, making difficulty-based selection impossible. For these domains, our findings suggest a simple and effective alternative: using the length of the CoT as a proxy for data selection. As demonstrated in Section \ref{sec:selection}, CoT length is the key driver of policy alignment. Distilling on longer CoT sequences not only enables the student model to maintain alignment stability over a long reasoning horizon, but also successfully activates advanced reasoning patterns (e.g. reflection), that may miss in short trajectories. 

\paragraph{Discussion on Hard Problems with Short Response.} 
A natural question arises regarding potential edge cases in our difficulty-driven selection strategy: what if a problem is categorized as hard due to low model accuracy, yet yields relatively short CoT reasoning responses? We term such instances pseudo-hard problems. These cases often stem from subtle trick questions, common calculation traps, or counterintuitive prompts where models frequently make premature errors without engaging in multi-step exploration. Because these prompts fail to elicit essential reasoning behaviors such as self-reflection, error correction, or backtracking, distilling on them may not yield superior performance gains. Nevertheless, our empirical observations indicate that such pseudo-hard examples are exceptionally rare in practice, as problem difficulty and generated response length exhibit a strong positive correlation. Consequently, while difficulty serves as a robust proxy for selection, incorporating response length as a secondary filtering metric could further refine data selection in edge scenarios.

\paragraph{The Limits of Scaling Rollout Horizon and the Exposure Bias Trade-off.}
A natural question is whether continuously increasing the maximum response length will always yield better policy alignment. We argue that there exists a empirical trade-off: while longer trajectories contain richer reasoning patterns, excessively long on-policy rollouts suffer from severe exposure bias. Because OPD optimizes the student policy against its own dynamically sampled trajectories, any reasoning errors made by the student model will compound over a long horizon. This compounding error drifts the student into out-of-distribution states, which significantly degrades the quality of the teacher's policy supervision. The maximum rollout length in our experiment is chosen as 7,168 following the empirical choice in \cite{li2026rethinking}. To train on even longer CoT sequences without policy decay, we think designing the appropriate training algorithm (e.g. Prune-OPD \citep{yang2026prune}) is necessary.

\section{Example Details} \label{app:sample_details}

\subsection{Accuracy of Sampled Examples}
To classify the difficulty of tasks in our example pool and select representative problems, we evaluate the baseline performance of both the student and teacher models. Specifically, we run 16 independent rollouts per problem for the student model (DeepSeek-R1-Distill-Qwen-1.5B) and the teacher model (JustRL-DeepSeek-1.5B), respectively. For each problem $i$, let $S_i$ and $T_i$ denote the rollout accuracies of the student and teacher models. We then compute the average accuracy as $A_i = (S_i + T_i)/2$ to represent the task difficulty.

Based on this metric, we rank all training examples and group them into three categories: (1) Easy ($A_i > 0.9$), (2) Medium ($0.1 \le A_i \le 0.9$), and (3) Hard ($A_i < 0.1$). Table \ref{tab:opd_accuracies} reports the detailed rollout accuracies of both models on the representative training problems selected in our experiments.

\begin{table}[htbp]
\centering
\caption{\textbf{Accuracies of Sampled Problems.} We report the rollout accuracies of both the student and teacher models across 16 rollouts per problem during our data preparation phase.}
\label{tab:opd_accuracies}
\small
\begin{tabular}{lcc|lcc|lcc}
\toprule
\multicolumn{3}{c}{\textbf{Easy Problems}} & \multicolumn{3}{c}{\textbf{Medium Problems}} & \multicolumn{3}{c}{\textbf{Hard Problems}} \\
\cmidrule(lr){1-3} \cmidrule(lr){4-6} \cmidrule(lr){7-9}
\textbf{ID} & \textbf{Student} & \textbf{Teacher} & \textbf{ID} & \textbf{Student} & \textbf{Teacher} & \textbf{ID} & \textbf{Student} & \textbf{Teacher} \\
\midrule
$\pi_{5}$  & 100.0 & 100.0 & $\pi_{316}$ & 68.8 & 100.0 & $\pi_{874}$ & 0.0 & 6.3  \\
$\pi_{20}$  & 100.0 & 100.0 & $\pi_{487}$ & 37.5 & 100.0 & $\pi_{890}$ & 0.0 & 0.0  \\
$\pi_{21}$  & 100.0 & 100.0 & $\pi_{530}$ & 25.0 & 100.0 & $\pi_{948}$ & 0.0 & 0.0  \\
$\pi_{63}$  & 93.8  & 100.0 & $\pi_{543}$ & 50.0 & 75.0  & $\pi_{954}$ & 0.0 & 0.0  \\
$\pi_{84}$  & 93.8  & 100.0 & $\pi_{661}$ & 25.0 & 68.8  & $\pi_{961}$ & 0.0 & 0.0  \\
$\pi_{105}$ & 87.5  & 100.0  & $\pi_{763}$ & 6.3  & 37.5  & $\pi_{973}$ & 0.0 & 0.0  \\
$\pi_{178}$ & 87.5  & 100.0 & $\pi_{794}$ & 0.0  & 37.5  & $\pi_{987}$ & 0.0 & 0.0  \\
$\pi_{192}$ & 87.5  & 100.0   & $\pi_{820}$ & 6.3  & 18.8  & $\pi_{997}$ & 0.0 & 0.0  \\
\bottomrule
\end{tabular}
\end{table}

\subsection{Content of Sampled Examples}

\begin{table}[htbp]
\centering
\caption{\textbf{Details of example} $\pi_{5}$.}
\begin{tabular}{p{0.95\textwidth}}
\toprule
\textbf{Prompt:} \\
\midrule
\footnotesize
\begin{lstlisting}[breaklines=true, basicstyle=\ttfamily, aboveskip=-1.0em, belowskip=0pt, breakindent=0pt]
Let $a$ and $b$ be positive real numbers.  Find the minimum value of
\[a^2 + b^2 + \frac{1}{(a + b)^2}.\]The answer is in the form k\sqrt{m}+n,. Please provide the value of k + m + n. Please reason step by step, and put your final answer within \boxed{{}}.
\end{lstlisting} \\
\midrule
\textbf{Ground truth:} $3$. \\
\bottomrule
\end{tabular}
\end{table}

\begin{table}[htbp]
\centering
\caption{\textbf{Details of example} $\pi_{20}$.}
\begin{tabular}{p{0.95\textwidth}}
\toprule
\textbf{Prompt:} \\
\midrule
\footnotesize
\begin{lstlisting}[breaklines=true, basicstyle=\ttfamily, aboveskip=-1.0em, belowskip=0pt, breakindent=0pt]
In trapezoid $ABCD$ the lengths of the bases $AB$ and $CD$ are 8 and 17 respectively. The legs of the trapezoid are extended beyond $A$ and $B$ to meet at point $E$. What is the ratio of the area of triangle $EAB$ to the area of trapezoid $ABCD$? Express your answer as a common fraction.The answer is in the form rac{m}{n}, where gcd(m, n) = 1. Please provide the value of m + n. Please reason step by step, and put your final answer within \boxed{{}}.
\end{lstlisting} \\
\midrule
\textbf{Ground truth:} $289$. \\
\bottomrule
\end{tabular}
\end{table}

\begin{table}[htbp]
\centering
\caption{\textbf{Details of example} $\pi_{21}$.}
\begin{tabular}{p{0.95\textwidth}}
\toprule
\textbf{Prompt:} \\
\midrule
\footnotesize
\begin{lstlisting}[breaklines=true, basicstyle=\ttfamily, aboveskip=-1.0em, belowskip=0pt, breakindent=0pt]
Lines $l_1^{}$ and $l_2^{}$ both pass through the origin and make first-quadrant angles of $\frac{\pi}{70}$ and $\frac{\pi}{54}$ radians, respectively, with the positive $x$-axis. For any line  $l$, the transformation $R(l)$ produces another line as follows: $l$ is reflected in $l_1$, and the resulting line is reflected in $l_2$. Let $R^{(1)}(l)=R(l)$ and $R^{(n)}(l)=R\left(R^{(n-1)}(l)\right)$. Given that $l$ is the line $y=\frac{19}{92}x$, find the smallest positive integer $m$ for which $R^{(m)}(l)=l$. Please reason step by step, and put your final answer within \boxed{{}}.
\end{lstlisting} \\
\midrule
\textbf{Ground truth:} $945$. \\
\bottomrule
\end{tabular}
\end{table}

\begin{table}[htbp]
\centering
\caption{\textbf{Details of example} $\pi_{63}$.}
\begin{tabular}{p{0.95\textwidth}}
\toprule
\textbf{Prompt:} \\
\midrule
\footnotesize
\begin{lstlisting}[breaklines=true, basicstyle=\ttfamily, aboveskip=-1.0em, belowskip=0pt, breakindent=0pt]
On the party, every boy gave $1$ candy to every girl, and every girl gave $1$ candy to every boy. Then every boy ate $2$ candies, and every girl ate $3$ candies. It is known that $\frac{1}{4}$ of all candies were eaten. Find the greatest possible number of children at the party. Please reason step by step, and put your final answer within \boxed{{}}.
\end{lstlisting} \\
\midrule
\textbf{Ground truth:} $35$. \\
\bottomrule
\end{tabular}
\end{table}

\begin{table}[htbp]
\centering
\caption{\textbf{Details of example} $\pi_{84}$.}
\begin{tabular}{p{0.95\textwidth}}
\toprule
\textbf{Prompt:} \\
\midrule
\footnotesize
\begin{lstlisting}[breaklines=true, basicstyle=\ttfamily, aboveskip=-1.0em, belowskip=0pt, breakindent=0pt]
The prime numbers $a$, $b$, and $c$ satisfy the equation $a + b^2 = 4c^2$. Determine the sum of all possible values of $a + b + c$. Please reason step by step, and put your final answer within \boxed{{}}.
\end{lstlisting} \\
\midrule
\textbf{Ground truth:} $31$. \\
\bottomrule
\end{tabular}
\end{table}

\begin{table}[htbp]
\centering
\caption{\textbf{Details of example} $\pi_{105}$.}
\begin{tabular}{p{0.95\textwidth}}
\toprule
\textbf{Prompt:} \\
\midrule
\footnotesize
\begin{lstlisting}[breaklines=true, basicstyle=\ttfamily, aboveskip=-1.0em, belowskip=0pt, breakindent=0pt]
Compute the smallest positive integer $N$ for which $N \cdot 2^{2024}$ is a multiple of $2024$. Please reason step by step, and put your final answer within \boxed{{}}.
\end{lstlisting} \\
\midrule
\textbf{Ground truth:} $253$. \\
\bottomrule
\end{tabular}
\end{table}

\begin{table}[htbp]
\centering
\caption{\textbf{Details of example} $\pi_{178}$.}
\begin{tabular}{p{0.95\textwidth}}
\toprule
\textbf{Prompt:} \\
\midrule
\footnotesize
\begin{lstlisting}[breaklines=true, basicstyle=\ttfamily, aboveskip=-1.0em, belowskip=0pt, breakindent=0pt]
Determine the number of pairs $(a,b)$ of real numbers such that $10, a, b, ab$ is an arithmetic progression. Please reason step by step, and put your final answer within \boxed{{}}.
\end{lstlisting} \\
\midrule
\textbf{Ground truth:} $2$. \\
\bottomrule
\end{tabular}
\end{table}

\begin{table}[htbp]
\centering
\caption{\textbf{Details of example} $\pi_{192}$.}
\begin{tabular}{p{0.95\textwidth}}
\toprule
\textbf{Prompt:} \\
\midrule
\footnotesize
\begin{lstlisting}[breaklines=true, basicstyle=\ttfamily, aboveskip=-1.0em, belowskip=0pt, breakindent=0pt]
Compute the remainder when $2^{3^5}+ 3^{5^2}+ 5^{2^3}$ is divided by $30$. Please reason step by step, and put your final answer within \boxed{{}}.
\end{lstlisting} \\
\midrule
\textbf{Ground truth:} $6$. \\
\bottomrule
\end{tabular}
\end{table}

\begin{table}[htbp]
\centering
\caption{\textbf{Details of example} $\pi_{316}$.}
\begin{tabular}{p{0.95\textwidth}}
\toprule
\textbf{Prompt:} \\
\midrule
\footnotesize
\begin{lstlisting}[breaklines=true, basicstyle=\ttfamily, aboveskip=-1.0em, belowskip=0pt, breakindent=0pt]
A dartboard consists of three concentric circles with radii 4, 6, and 8. Three darts are thrown at the board, sticking at random locations. Determine the probability that each dart lands in a different region of the dartboard. The probability can be expressed as \( \frac{m}{n} \), where \( m \) and \( n \) are relatively prime positive integers. Calculate \( m + n \). Please reason step by step, and put your final answer within \boxed{{}}.
\end{lstlisting} \\
\midrule
\textbf{Ground truth:} $617$. \\
\bottomrule
\end{tabular}
\end{table}

\begin{table}[htbp]
\centering
\caption{\textbf{Details of example} $\pi_{487}$.}
\begin{tabular}{p{0.95\textwidth}}
\toprule
\textbf{Prompt:} \\
\midrule
\footnotesize
\begin{lstlisting}[breaklines=true, basicstyle=\ttfamily, aboveskip=-1.0em, belowskip=0pt, breakindent=0pt]
In how many ways can the integers from 1 to n be ordered subject to the condition that, except for the first integer on the left, every integer differs by 1 from some integer to the left of it? Please provide the number of ways for $n = 6$. Please reason step by step, and put your final answer within \boxed{{}}.
\end{lstlisting} \\
\midrule
\textbf{Ground truth:} $32$. \\
\bottomrule
\end{tabular}
\end{table}

\begin{table}[htbp]
\centering
\caption{\textbf{Details of example} $\pi_{530}$.}
\begin{tabular}{p{0.95\textwidth}}
\toprule
\textbf{Prompt:} \\
\midrule
\footnotesize
\begin{lstlisting}[breaklines=true, basicstyle=\ttfamily, aboveskip=-1.0em, belowskip=0pt, breakindent=0pt]
Determine the largest integer $N$ for which there exists a $6 \times N$ table $T$ that has the following properties:\n\n- Every column contains the numbers $1, 2, \ldots, 6$ in some ordering.\n- For any two columns $i \ne j$, there exists a row $r$ such that $T(r,i) = T(r,j)$.\n- For any two columns $i \ne j$, there exists a row $s$ such that $T(s,i) \ne T(s,j)$. Please reason step by step, and put your final answer within \boxed{{}}.
\end{lstlisting} \\
\midrule
\textbf{Ground truth:} $120$. \\
\bottomrule
\end{tabular}
\end{table}

\begin{table}[htbp]
\centering
\caption{\textbf{Details of example} $\pi_{543}$.}
\begin{tabular}{p{0.95\textwidth}}
\toprule
\textbf{Prompt:} \\
\midrule
\footnotesize
\begin{lstlisting}[breaklines=true, basicstyle=\ttfamily, aboveskip=-1.0em, belowskip=0pt, breakindent=0pt]
Let f be a polynomial of degree $3$ with integer coefficients such that $f(0) = 3$ and $f(1) = 11$.
If f has exactly $2$ integer roots, how many such polynomials $f$ exist? Please reason step by step, and put your final answer within \boxed{{}}.
\end{lstlisting} \\
\midrule
\textbf{Ground truth:} $0$. \\
\bottomrule
\end{tabular}
\end{table}

\begin{table}[htbp]
\centering
\caption{\textbf{Details of example} $\pi_{661}$.}
\begin{tabular}{p{0.95\textwidth}}
\toprule
\textbf{Prompt:} \\
\midrule
\footnotesize
\begin{lstlisting}[breaklines=true, basicstyle=\ttfamily, aboveskip=-1.0em, belowskip=0pt, breakindent=0pt]
Find the area of the figure on the coordinate plane bounded by the straight lines $x = 0$, $x = 2$, and the graphs of the functions $y = \\sqrt{x^3 + 1}$ and $y = -\\sqrt[3]{x^2 + 2x}$. Please reason step by step, and put your final answer within \\boxed{{}}.
\end{lstlisting} \\
\midrule
\textbf{Ground truth:} $6$. \\
\bottomrule
\end{tabular}
\end{table}

\begin{table}[htbp]
\centering
\caption{\textbf{Details of example} $\pi_{763}$.}
\begin{tabular}{p{0.95\textwidth}}
\toprule
\textbf{Prompt:} \\
\midrule
\footnotesize
\begin{lstlisting}[breaklines=true, basicstyle=\ttfamily, aboveskip=-1.0em, belowskip=0pt, breakindent=0pt]
An organization has $30$ employees, $20$ of whom have a brand A computer while the other $10$ have a brand B computer. For security, the computers can only be connected to each other and only by cables. The cables can only connect a brand A computer to a brand B computer. Employees can communicate with each other if their computers are directly connected by a cable or by relaying messages through a series of connected computers. Initially, no computer is connected to any other. A technician arbitrarily selects one computer of each brand and installs a cable between them, provided there is not already a cable between that pair. The technician stops once every employee can communicate with each other. What is the maximum possible number of cables used? Please reason step by step, and put your final answer within \\boxed{{}}.
\end{lstlisting} \\
\midrule
\textbf{Ground truth:} $191$. \\
\bottomrule
\end{tabular}
\end{table}

\begin{table}[htbp]
\centering
\caption{\textbf{Details of example} $\pi_{794}$.}
\begin{tabular}{p{0.95\textwidth}}
\toprule
\textbf{Prompt:} \\
\midrule
\footnotesize
\begin{lstlisting}[breaklines=true, basicstyle=\ttfamily, aboveskip=-1.0em, belowskip=0pt, breakindent=0pt]
Tanya wrote numbers in the form $n^7 - 1$ for $n = 2, 3, \ldots$ and noticed that for $n = 8$, she obtained a number divisible by $337$. For what minimal $n$ did she get a number divisible by $2022$? Please reason step by step, and put your final answer within \boxed{{}}.
\end{lstlisting} \\
\midrule
\textbf{Ground truth:} $79$. \\
\bottomrule
\end{tabular}
\end{table}

\begin{table}[htbp]
\centering
\caption{\textbf{Details of example} $\pi_{820}$.}
\begin{tabular}{p{0.95\textwidth}}
\toprule
\textbf{Prompt:} \\
\midrule
\footnotesize
\begin{lstlisting}[breaklines=true, basicstyle=\ttfamily, aboveskip=-1.0em, belowskip=0pt, breakindent=0pt]
Vasya has $n$ candies of several types, where $n > 145$. It is known that for any group of at least 145 candies, there is a type of candy which appears exactly 10 times. Find the largest possible value of $n$. Please reason step by step, and put your final answer within \boxed{{}}.
\end{lstlisting} \\
\midrule
\textbf{Ground truth:} $160$. \\
\bottomrule
\end{tabular}
\end{table}

\begin{table}[htbp]
\centering
\caption{\textbf{Details of example} $\pi_{874}$.}
\begin{tabular}{p{0.95\textwidth}}
\toprule
\textbf{Prompt:} \\
\midrule
\footnotesize
\begin{lstlisting}[breaklines=true, basicstyle=\ttfamily, aboveskip=-1.0em, belowskip=0pt, breakindent=0pt]
Find the number of subsets of $\{1,3,5,7,9,11,13,15,17,19\}$ where the elements in the subset add to $49$. Please reason step by step, and put your final answer within \boxed{{}}.
\end{lstlisting} \\
\midrule
\textbf{Ground truth:} $22$. \\
\bottomrule
\end{tabular}
\end{table}

\begin{table}[htbp]
\centering
\caption{\textbf{Details of example} $\pi_{890}$.}
\begin{tabular}{p{0.95\textwidth}}
\toprule
\textbf{Prompt:} \\
\midrule
\footnotesize
\begin{lstlisting}[breaklines=true, basicstyle=\ttfamily, aboveskip=-1.0em, belowskip=0pt, breakindent=0pt]
Let $x_1, x_2, \dots, x_{100}$ be real numbers such that $|x_1| = 63$ and $|x_{n+1}| = |x_n + 1|$ for $n = 1, 2, \dots, 99$. Find the largest possible value of $(-x_1 - x_2 - \cdots - x_{100})$. Please reason step by step, and put your final answer within \boxed{{}}.
\end{lstlisting} \\
\midrule
\textbf{Ground truth:} $2034$. \\
\bottomrule
\end{tabular}
\end{table}

\begin{table}[htbp]
\centering
\caption{\textbf{Details of example} $\pi_{948}$.}
\begin{tabular}{p{0.95\textwidth}}
\toprule
\textbf{Prompt:} \\
\midrule
\footnotesize
\begin{lstlisting}[breaklines=true, basicstyle=\ttfamily, aboveskip=-1.0em, belowskip=0pt, breakindent=0pt]
Two real numbers $x$ and $y$ are chosen at random in the interval (0,1) with respect to the uniform distribution. What is the probability that the closest integer to $x/y$ is even? The original answer is in the form $r + s\pi$, please give the value of $r + s$.  Please reason step by step, and put your final answer within \boxed{{}}.
\end{lstlisting} \\
\midrule
\textbf{Ground truth:} $4$. \\
\bottomrule
\end{tabular}
\end{table}

\begin{table}[htbp]
\centering
\caption{\textbf{Details of example} $\pi_{954}$.}
\begin{tabular}{p{0.95\textwidth}}
\toprule
\textbf{Prompt:} \\
\midrule
\footnotesize
\begin{lstlisting}[breaklines=true, basicstyle=\ttfamily, aboveskip=-1.0em, belowskip=0pt, breakindent=0pt]
Let $f:\{1,2,\dots,2019\}\to\{-1,1\}$ be a function, such that for every $k\in\{1,2,\dots,2019\}$, there exists an $\ell\in\{1,2,\dots,2019\}$ such that $$ \sum_{i\in\mathbb{Z}:(\ell-i)(i-k)\geqslant 0} f(i)\leqslant 0. $$ Determine the maximum possible value of $$ \sum_{i\in\mathbb{Z}:1\leqslant i\leqslant 2019} f(i). $$ Please reason step by step, and put your final answer within \boxed{{}}.
\end{lstlisting} \\
\midrule
\textbf{Ground truth:} $673$. \\
\bottomrule
\end{tabular}
\end{table}

\begin{table}[htbp]
\centering
\caption{\textbf{Details of example} $\pi_{961}$.}
\begin{tabular}{p{0.95\textwidth}}
\toprule
\textbf{Prompt:} \\
\midrule
\footnotesize
\begin{lstlisting}[breaklines=true, basicstyle=\ttfamily, aboveskip=-1.0em, belowskip=0pt, breakindent=0pt]
A herder has forgotten the number of cows she has and does not want to count all of them. She remembers these four facts about the number of cows:\n\n- It has $3$ digits.\n- It is a palindrome.\n- The middle digit is a multiple of $4$.\n- It is divisible by $11$.\n\nWhat is the sum of all possible numbers of cows that the herder has?\n\n$\\textbf{(A) }343 \\ \\textbf{(B) }494 \\ \\textbf{(C) }615 \\ \\textbf{(D) }635 \\ \\textbf{(E) }726$ Please reason step by step, and put your final answer within \\boxed{{}}.
\end{lstlisting} \\
\midrule
\textbf{Ground truth:} $726$. \\
\bottomrule
\end{tabular}
\end{table}

\begin{table}[!t]
\centering
\caption{\textbf{Details of example} $\pi_{973}$.}
\begin{tabular}{p{0.95\textwidth}}
\toprule
\textbf{Prompt:} \\
\midrule
\footnotesize
\begin{lstlisting}[breaklines=true, basicstyle=\ttfamily, aboveskip=-1.0em, belowskip=0pt, breakindent=0pt]
Find all ordered pairs $(a, b)$ of positive integers for which\n$$\n\frac{1}{a}+\frac{1}{b}=\frac{3}{2018} .\n$$\nPlease provide the sum of all integers in the ordered pairs. Please reason step by step, and put your final answer within \boxed{{}}.
\end{lstlisting} \\
\midrule
\textbf{Ground truth:} $1438383$. \\
\bottomrule
\end{tabular}

\caption{\textbf{Details of example} $\pi_{987}$.}
\begin{tabular}{p{0.95\textwidth}}
\toprule
\textbf{Prompt:} \\
\midrule
\footnotesize
\begin{lstlisting}[breaklines=true, basicstyle=\ttfamily, aboveskip=-1.0em, belowskip=0pt, breakindent=0pt]
Find the smallest positive integer $n$ with the following property:  for every sequence of positive integers $a_1,a_2,\ldots , a_n$ with $a_1+a_2+\ldots +a_n=2013$, there exist some (possibly one) consecutive term(s) in the sequence that add up to $70$. Please reason step by step, and put your final answer within \boxed{{}}.
\end{lstlisting} \\
\midrule
\textbf{Ground truth:} $1033$. \\
\bottomrule
\end{tabular}

\caption{\textbf{Details of example} $\pi_{997}$.}
\begin{tabular}{p{0.95\textwidth}}
\toprule
\textbf{Prompt:} \\
\midrule
\footnotesize
\begin{lstlisting}[breaklines=true, basicstyle=\ttfamily, aboveskip=-1.0em, belowskip=0pt, breakindent=0pt]
A quadratic polynomial $f(x)$ is called sparse if its degree is exactly 2 , if it has integer coefficients, and if there exists a nonzero polynomial $g(x)$ with integer coefficients such that $f(x) g(x)$ has degree at most 3 and $f(x) g(x)$ has at most two nonzero coefficients. Find the number of sparse quadratics whose coefficients lie between 0 and 10, inclusive. Please reason step by step, and put your final answer within \boxed{{}}.
\end{lstlisting} \\
\midrule
\textbf{Ground truth:} $228$. \\
\bottomrule
\end{tabular}
\end{table}

\end{document}